\documentclass[11pt]{article}
\usepackage[preprint]{acl}

\usepackage[english,bidi=default]{babel}
\babelfont{rm}{TeXGyreTermesX}

\babelprovide[import]{russian}
\babelprovide[import]{bulgarian}
\babelfont[russian]{rm}[Path=fonts/, BoldFont=FreeSerifBold.otf]{FreeSerif.otf} 

\babelprovide[import]{persian}
\babelfont[*arabic]{rm}[Path=fonts/]{NotoNaskhArabic-Regular.ttf}

\usepackage{fontawesome5}
\usepackage{twemojis}
\usepackage{microtype}
\usepackage{inconsolata}
\usepackage{graphicx}
\usepackage{examples-slim}
\usepackage{amsmath}
\usepackage{amssymb}
\usepackage{amsfonts}
\usepackage{todonotes}
\usepackage{booktabs}
\usepackage{latexsym}
\usepackage{subcaption}
\usepackage{tabularx}
\usepackage{tabulary}
\usepackage{float}
\usepackage{stfloats}
\usepackage[table]{xcolor}
\usepackage{fontspec}
\usepackage{pifont}
\newcommand{\yes}{\textcolor{green!50!black}{\ding{51}}}
\newcommand{\no}{\textcolor{red}{\ding{55}}}

\newcommand{\cy}[1]{\foreignlanguage{russian}{#1}}
\newcommand{\ar}[1]{\foreignlanguage{persian}{#1}}
\newfontfamily\zhfont[Path=fonts/]{NotoSerifCJKsc-Regular.otf}[Scale=0.7]
\newfontfamily\jpfont[Path=fonts/]{NotoSerifJP-Regular.otf}[Scale=0.7]
\newfontfamily\krfont[Path=fonts/]{NotoSerifKR-Regular.otf}[Scale=0.7]

\newcommand{\zh}[1]{{\zhfont #1}}
\newcommand{\jp}[1]{{\jpfont #1}}
\newcommand{\ko}[1]{{\krfont #1}}

\usepackage{cleveref}
\crefformat{section}{\S#2#1#3}
\crefformat{subsection}{\S#2#1#3}
\crefformat{subsubsection}{\S#2#1#3}
\crefformat{paragraph}{\P#2#1#3}
\crefformat{subparagraph}{\P#2#1#3}
\crefmultiformat{section}{\S#2#1#3}{ and~\S#2#1#3}{, \S#2#1#3}{, and~\S#2#1#3}
\crefmultiformat{subsection}{\S#2#1#3}{ and~\S#2#1#3}{, \S#2#1#3}{, and~\S#2#1#3}
\crefmultiformat{subsubsection}{\S#2#1#3}{ and~\S#2#1#3}{, \S#2#1#3}{, and~\S#2#1#3}
\crefmultiformat{paragraph}{\P\P#2#1#3}{ and~#2#1#3}{, #2#1#3}{, and~#2#1#3}
\crefmultiformat{subparagraph}{\P\P#2#1#3}{ and~#2#1#3}{, #2#1#3}{, and~#2#1#3}
\crefrangeformat{section}{\mbox{\S\S#3#1#4--#5#2#6}}
\crefrangeformat{subsection}{\mbox{\S\S#3#1#4--#5#2#6}}
\crefrangeformat{subsubsection}{\mbox{\S\S#3#1#4--#5#2#6}}
\crefrangeformat{paragraph}{\mbox{\P\P#3#1#4--#5#2#6}}
\crefrangeformat{subparagraph}{\mbox{\P\P#3#1#4--#5#2#6}}
\crefname{part}{Part}{Parts}
\Crefname{part}{Part}{Parts}
\crefname{chapter}{Ch.}{Ch.}
\Crefname{chapter}{Ch.}{Ch.}
\crefname{footnote}{Fn.}{Fn.}
\Crefname{footnote}{Fn.}{Fn.}
\crefname{figure}{Figure}{Figures}
\crefname{table}{Table}{Tables}
\crefname{subfigure}{Figure}{Figures}
\Crefname{subfigure}{Figure}{Figures}
\crefname{algocf}{Algorithm}{Algorithms}
\Crefname{algocf}{Algorithm}{Algorithms}
\crefname{xnumi}{ex.}{exs.}
\Crefname{xnumi}{Ex.}{Exs.}
\crefname{xnumii}{ex.}{exs.}
\Crefname{xnumii}{Ex.}{Exs.}

\crefname{appendix}{\S}{\S\S}
\Crefname{appendix}{\S}{\S\S}
\crefformat{appendix}{\S#2#1#3}
\Crefformat{appendix}{\S#2#1#3}
\crefrangeformat{appendix}{\mbox{\S\S#3#1#4--#5#2#6}}
\Crefrangeformat{appendix}{\mbox{\S\S#3#1#4--#5#2#6}}

\newcommand{\maxodds}{\textbf{\textsc{Max Odds}}}
\newcommand{\odds}{\textbf{\textsc{Odds}}}

\usepackage{multirow}
\usepackage{makecell}
\usepackage{fontspec}
\usepackage{hyperref}
\title{Causal Interventions Reveal Typologically Organized\\Syntactic Mechanisms in Multilingual Language Models}

\author{
  Sasha Boguraev\textsuperscript{1} \quad
  Toshiki Nakai\textsuperscript{2,}\thanks{Work initiated while author was at Saarland University.} \quad
  Kyle Mahowald\textsuperscript{1} \quad
  Julius Steuer\textsuperscript{3} \\
  \textsuperscript{1}The University of Texas at Austin \quad
  \textsuperscript{2}Leipzig University, ScaDS.AI Dresden/Leipzig \\
  \textsuperscript{3}Heidelberg Institute for Theoretical Studies \\
  \texttt{\{sasha.boguraev, kyle\}@utexas.edu} \\ \texttt{toshiki.nakai@uni-leipzig.de} \quad \texttt{julius.steuer@h-its.org}
}

\begin{document}
\maketitle
\begin{abstract}
Linguistic theory has long recognized cross-linguistic syntactic regularities, leading to claims that these similar structures are processed by similar mechanisms. However, this hypothesis has been difficult to test empirically due to our lack of fine-grained, manipulable access of human processing mechanisms. In this work, we take advantage of techniques from mechanistic interpretability to study such a question in multilingual LMs. We first isolate language-internal mechanisms before attempting to transfer them cross-lingually. Across four models and three well-studied constructions (subject--verb number agreement, anaphoric pronoun gender agreement, and filler--gap object extraction) we find consistent cross-lingual mechanism transfer. We further find transfer to be graded, with more transfer between more typologically similar languages. We believe our work provides novel hypotheses about cross-linguistic syntactic structures and multilingual processing, and more broadly shows how the study of language models can help inform linguistic theory.
\end{abstract}

\begin{figure}[ht!]
    \centering
    \includegraphics[width=0.9\linewidth]{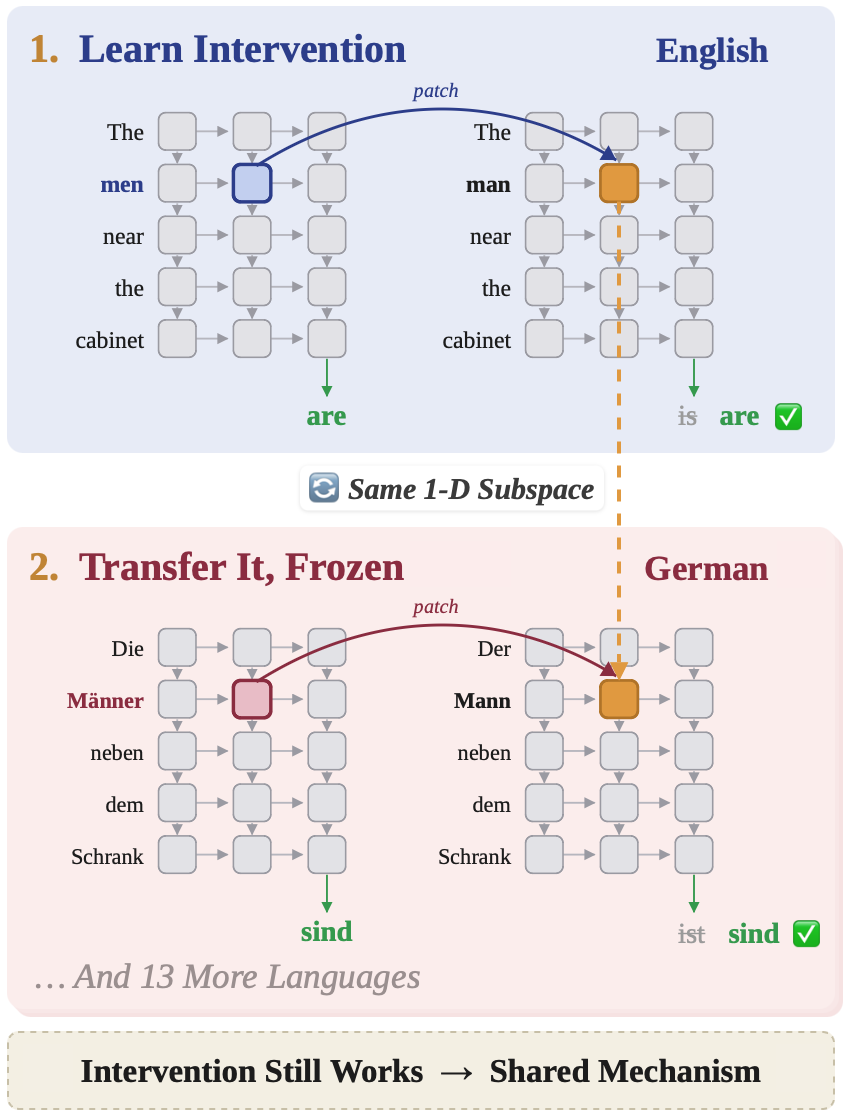}
    \caption{We localize LM's syntactic mechanisms to language-internal subspaces. We then test such mechanism's effects on other languages. If the intervention is still efficacious, this indicates the model's syntactic mechanisms are language-agnostic. This provides evidence of \emph{cross-lingual shared mechanisms}. 
    }
    \label{fig:fig1}
    \vspace{-10pt}
\end{figure}

\section{Introduction}
Linguistics has long accepted there are shared abstract syntactic structures across different languages --- perhaps due to innate constraints \citep{Chomsky1965} or
functional evolutionary pressures \citep{greenberg1963some,comrie1989language}.
It has often been hypothesized that these similar structures across different languages recruit similar machinery in human processing \citep{hartsuiker2004syntax,a2014cross,norcliffe2015cross,malik2022investigation}.
However, testing for such shared mechanisms in humans is difficult due to a lack of granular manipulable access needed to ask whether processing of two languages recruit the \emph{same} internal mechanism. 

In recent years, neural Language Models (LMs) have shown remarkable syntactic competence --- possessing an ability to produce and process utterances previously thought to require abstract linguistic representations, both mono- and cross-lingually \citep{linzen-etal-2016-assessing,wilcox-etal-2018-rnn,manning2020emergent,hu-etal-2020-systematic,warstadt-etal-2020-blimp-benchmark,jumelet-etal-2026-multiblimp}. This competence, combined with privileged access to their internal computations has led to a host of work treating them as model organisms to formulate and test linguistic hypotheses \citep{linzen2021syntactic, warstadt2022artificial, wilcox2024using,boguraev-etal-2025-causal, boguraev2026causaldrawbridgescharacterizinggradient, misra2026systematicframeworkgeneratingnovel}.

In this work, we take advantage of advances in mechanistic interpretability, particularly Causal Abstraction \citep{geiger2025causal}, to understand the internal organization of high-level linguistic abstractions in multilingual LMs. 
The basic idea of the Causal Abstraction framework is to identify internal structure in the neural model that controls downstream behavior: for our purposes, syntactic behavior \citep{arora_causalgym_2024,boguraev-etal-2025-causal}.
Our goal here is to identify causally relevant linguistic structures in one language and then test whether intervening on that same structure in \textit{another} language has a similar effect (\Cref{fig:fig1}).
If so, we would take that to be evidence for shared mechanisms across languages---offering potential insights into the underlying similarity of structures across languages, as well as into how multilingual neural models organize syntactic information.

In particular, we test for shared mechanisms across three structurally varied phenomena:
subject--verb number agreement, anaphoric pronoun gender agreement, and filler--gap object extraction. Across fifteen languages and four LMs spanning sizes, families, and training specs, we find LMs recruit shared, repurposed mechanisms for each phenomenon rather than language-specific ones. We further provide evidence that this representational sharing is driven by typological similarity, with transfer organized around typological `hubs' --- languages broadly similar to a large amount of the other studied languages. 

Taken together, our results characterize multilingual syntactic competence in LMs as the result of abstractions repurposed across languages. We further believe our results provide novel hypotheses about multilingual linguistic organization which could potentially be tested in humans.

\section{Cross-Lingual LM Interpretability}

When multilingual LMs with syntactic capabilities in multiple languages emerged, there was a flurry of interest in whether they learned shared representations across languages or whether they mostly learned each language independently.
Early work in this space took advantage of probing --- i.e., training classifiers on the internal states of an LM to recover high-level labels. \citet{chi-etal-2020-finding} used structural probes \citep{hewitt-manning-2019-structural} to show universal-dependency relations could be localized to language-agnostic subspaces, suggesting linguistic representations pattern similarly cross-lingually and \citet{papadimitriou-etal-2021-deep} demonstrated probes trained to recover high-level semantic features like `subjecthood' transferred in linguistically notable ways across languages. Such results were evidence of language-agnostic linguistic features in LMs. 

However, probing can be over-expressive \citep{hewitt-liang-2019-designing,voita-titov-2020-information}, prompting a shift to \emph{causal methods} which ensure faithfulness by manipulating internal states and checking for behavioral effect. Early iterations of these methods produced linguistic insights, finding both language-neutral and language-specific representations \citep{gonen-etal-2020-greek, srinivasan-etal-2023-counterfactually}, and finding cross-lingual grammatical circuits in English and Spanish \citep{ferrando-costa-jussa-2024-similarity}.  
However, such variants could not elucidate the `features' utilized for computation, merely localizing behavior to components. 

To surface such features, sparse-decomposition methods such as Sparse Auto-Encoders (SAEs) and Cross-Layer Transcoders (CLTs) were adopted. These methods likewise found cross-lingually shared feature-circuits \citep{brinkmann-etal-2025-large} and that language identity is encoded in later model layers \citep{harasse_2026_tracing_multilingual_representations}.
However, these methods find `interpretable features' in untrained models \citep{heap2025sparse} and are unstable across training runs \citep{paulo2025sparse, leask2025sparse}.

We instead use Distributed Alignment Search \citep[DAS;][]{geiger_finding_2024}, to localize abstract features in multilingual LMs. DAS is highly efficacious for syntactic interpretability \citep{arora_causalgym_2024}, and has been utilized to study the representational organization of syntactic features in English filler--gap constructions \citep{boguraev-etal-2025-causal} and their respective island constraints \citep{boguraev2026causaldrawbridgescharacterizinggradient}. 
Our methodology allows precise tests of whether causally implicated syntactic mechanisms transfer cross-lingually.

\section{Methods}

\subsection{Data}\label{sec:methods}

\begin{table*}[t]
\footnotesize
\centering
\setlength{\tabcolsep}{6pt}
\renewcommand{\arraystretch}{1.1}
\newcolumntype{W}{>{\hsize=1.5\hsize}X}
\newcolumntype{n}{>{\hsize=0.875\hsize}X}
\begin{tabularx}{\textwidth}{@{}p{3.1cm} n W n n n@{}}
\toprule
\rowcolor{gray!30}\multicolumn{2}{@{}l}{\textit{\textbf{Subject--Verb Number Agreement}}} & \textsc{NP$_1$} & \textsc{PP} & \textsc{NP$_2$} & \textsc{VP} \\
\textsc{English}   & & \textbf{The man / The men} & near & the cabinet & \textbf{is / are} \\
\textsc{Italian}   & & \textbf{L'uomo / Gli uomini} & vicino alla & tavola & \textbf{è / sono} \\
\textsc{Bulgarian} & & \cy{\textbf{Мъжът / Мъжете}} & \cy{до} & \cy{масата} & \cy{\textbf{е / са}} \\
\midrule
\rowcolor{gray!30}\multicolumn{2}{@{}l}{\textit{\textbf{Anaphoric Pronoun Gender Agreement}}} & \textsc{Name} & \textsc{VP} & \textsc{because} & \textsc{Pron.} \\
\textsc{English}   & & \textbf{James / Mary} & lied & because & \textbf{he / she} \\
\textsc{Italian}   & & \textbf{Mario / Maria} & rise & perché & \textbf{lui / lei} \\
\textsc{Bulgarian} & & \cy{\textbf{Иван / Мария}} & \cy{вика} & \cy{защото} & \cy{\textbf{той / тя}} \\
\midrule
\rowcolor{gray!30}\textit{\textbf{Filler--Gap (Object)}} & \textsc{Prefix} & \textsc{Filler} & \textsc{NP} & \textsc{VP} & \textsc{Gap} \\
\textsc{English}   & I know & \textbf{that / what} & she & saw & \textbf{the / .} \\
\textsc{Italian}   & So & \textbf{che / cosa} & lei & vide & \textbf{il / .} \\
\textsc{Bulgarian} & \cy{Знам,} & \cy{\textbf{че / какво}} & \cy{тя} & \cy{видя} & \cy{\textbf{него / .}} \\
\bottomrule
\end{tabularx}
\caption{Exemplar minimal pairs for a representative subset of languages with a full set of exemplars are in \Cref{appsec:templates}. Our gender agreement stimuli are violable: \textit{``Mary was powerful. James lied because \textbf{she} told him to''}, although we do not find this to affect our LMs behaviorally. Instead, we find our studied LMs to score at or above 90\% accuracy on 90\% (115/128) of templates and below 80\% accuracy only once.}
\label{tab:main-examples}
\vspace{-5pt}
\end{table*}

\paragraph{Stimuli} We study three constructions --- subject--verb number agreement, anaphoric pronoun gender agreement, and object extraction from an embedded \textit{wh}-question --- across fifteen languages --- English, French, German, Spanish, Italian, Portuguese, Russian, Bulgarian, Farsi, Dutch, Chinese, Japanese, Korean, Galician and Romanian. We design data templates as in \citet{arora_causalgym_2024} to allow sampling of large sets of unique minimal pairs (examples in \Cref{tab:main-examples}; full templates in \Cref{appsec:templates}).\footnote{As German and Dutch place embedded verbs clause-finally, their filler--gap templates cannot be perfectly aligned; we therefore exclude the \textsc{VP} position for these languages and analyze it only across the five with a non-final embedded verb.} Typological constraints prevent every language from being represented in each construction (e.g., Chinese does not have grammatical number). Nevertheless, there is a core set of seven languages across templates: English, German, Italian, Russian, Bulgarian, Dutch, and French. Behavioral evaluation shows LMs correctly process the stimuli with details in 
\Cref{appsec:behavior}.

\paragraph{Controls} We design three controls. \textsc{Random Labels} measures the specificity of our interventions. Per \citet{hewitt-liang-2019-designing} this control keeps the critical context but pairs it with unrelated, ungrammatical labels. An exemplar is in (1). 

{\exampleindent=2em
\begin{examples}
\item The \textbf{man/men} beside the car $\rightarrow$ \textbf{dog/run}
\end{examples}}

\noindent Interventions demonstrating strong specificity, as desired, \emph{should not generalize} to this control. \textsc{Other phenomena} evaluates each core-language intervention on the other constructions. We again \emph{expect no transfer}, as the constructions are syntactically distinct. Finally, following \citet{kumon-yanaka-2026-fine}, \textsc{OOD labels} consists of templates with the same critical context but with novel grammatical labels. We cannot extend this control to gender agreement, as there are no cross-lingually robust alternate labels. As such, we only report related results in \Cref{appsec:ood_evals}.

\subsection{Models}

We study four models: \texttt{mGPT} \citep[1.3b params.;][]{shliazhko2024mgpt} and \texttt{tiny-aya-base} \citep[3.4b params.;][]{salamanca2026tinyayabridgingscale} both expressly multilingual, trained on 61 and 70+ languages respectively, as well as \texttt{Llama-3.2-1b} and  \texttt{Llama-3.2-3b} \citep{grattafiori2024llama3herdmodels}, both not expressly multilingual, trained on generic web data, but nevertheless possessing multilingual capabilities. By evaluating four models, we can study effects of architectures and training factors (e.g. model size, training data composition) on representation sharing and ensure our findings reflect general trends rather than that of one model.

\subsection{Distributed Alignment Search}

DAS is a supervised interpretability method which finds low-rank $d$-dimensional subspaces in an LM in which a high-level causal variable can be localized and manipulated. Specifically, given a base value $b\in \mathbb{R}^n$ --- a tensor at a specific internal site when an LM processes some input --- and corresponding source value $s\in\mathbb{R}^n$ --- the tensor at the same site when the model processes a minimally-paired input --- DAS finds a subspace in which intervening from $s\rightarrow b$, whilst keeping all orthogonal components fixed, changes the output prediction correspondingly. This is operationalized as

\begin{equation*}
    \textbf{b} + (\textbf{sa}^\top - \textbf{ba}^\top)\textbf{a}
\end{equation*}

\noindent where $a\in\mathbb{R}^{n\times d}$ is a rotation matrix learned by minimizing the cross-entropy loss of the LM's prediction under intervention through gradient descent. Following \citet{arora_causalgym_2024, boguraev-etal-2025-causal, boguraev2026causaldrawbridgescharacterizinggradient} we learn 1D subspaces ($d=1$) in the residual stream of our given LMs. When the span we are intervening on contains multiple tokens, we intervene on the pooled representation across the span, and when two separate template positions combine into one token (i.e., French elision producing the single token \textit{qu'il} from \textit{que} $+$ \textit{il}) we evaluate both corresponding interventions at the combined token.

\paragraph{Training and Evaluation}

We train interventions at each template position and model layer. Following \citet{arora_causalgym_2024, boguraev-etal-2025-causal}, we train interventions for 100 steps with a batch size of 4, filtering training items to those on which the LM is behaviorally competent. We use $\odds$ to evaluate the interventions. This metric measures the probability increase for the source label after intervention, relative to the probability decrease of the base label. Higher $\odds$ indicate higher causal efficacy. In cases of aggregation, we report the $\maxodds$: the maximum $\odds$ value across layers at a given position. The use of these metrics is consistent with other work using DAS for linguistic evaluation \citep{arora_causalgym_2024, boguraev-etal-2025-causal, boguraev2026causaldrawbridgescharacterizinggradient}. We measure these metrics across a held-out evaluation set of 100 unique minimal pairs.

\section{Experiment 1: Do LMs Share Mechanisms Cross-Lingually?}\label{sec:exp1}

In our first experiment, we ask whether the mechanisms LMs use to process a given construction transfer cross-lingually.
That is, we use DAS to find a subspace causally implicated in a particular syntactic phenomenon (e.g., subject--verb agreement) in Language A and see if that same subspace can causally control the same phenomenon in Language B. In principle, this subspace could exploit the lexically-matched templates to discover translation-equivalent alignments \citep[à la][]{gonen-etal-2020-greek}, rather than abstract syntactic features. We flag this alternative hypothesis here and return to distinguishing it from genuine syntactic abstraction in the discussions of \Cref{sec:exp1} and \Cref{sec:exp2}.

\paragraph{Setup} We measure the $\odds$ across layers and positions for all learned interventions evaluated across the same construction multilingually, as well as on controls. We report the performance across distinct groups: \textbf{within-language interventions} --- e.g., training on English subject--verb agreement and evaluating on English subject--verb agreement, \textbf{cross-lingual interventions} --- e.g., training on English subject--verb agreement and evaluating on subject--verb agreement in all other languages, \textbf{within-family interventions} --- e.g., training on English subject--verb agreement in English and evaluating on subject--verb agreement of all other Germanic languages, and \textbf{controls} --- \textsc{random labels} and \textsc{other phenomena} in the main text, with \textsc{OOD labels} in \Cref{appsec:ood_evals}.

We compare these values across layers, positions, and models. We also run statistical tests on the $\maxodds$ of transferred interventions normalized by within-language performance. This intuitively measures how much an LM's mechanisms for a given language generalize to other languages with respect to its own, baseline, performance.

\paragraph{Hypothesis} We hypothesize there are shared syntactic mechanisms in LMs. That is, we expect cross-lingual transfer to be significantly above controls. We further expect higher transfer within a family than to the full set of evaluated languages.

\begin{figure*}
    \centering
    \includegraphics[width=\linewidth]{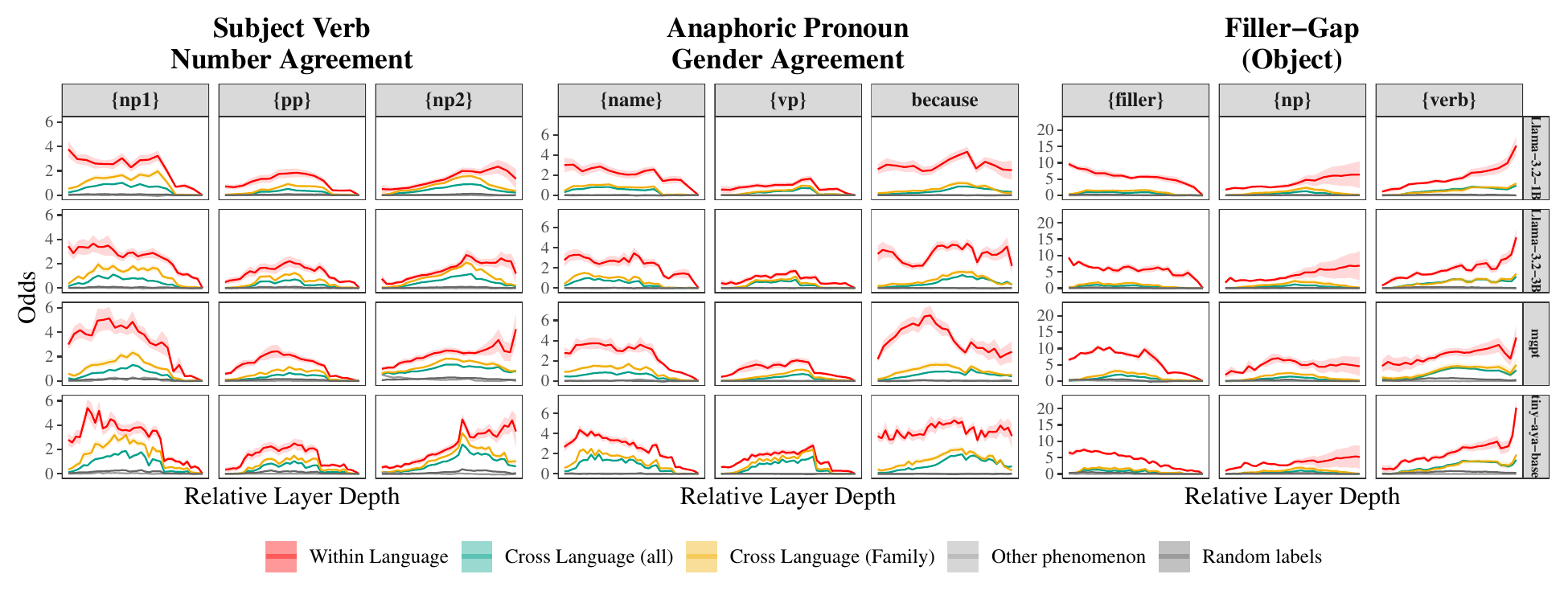}
    \caption{We calculate the \textbf{\textsc{Odds}} when evaluating our trained interventions on a held-out test set of the same language they are trained on (red line), all other languages (green line), all other languages within the same family (yellow line), and across two controls (grey and black lines). We find that across all positions, phenomena, and models, interventions consistently generalize across languages with pairwise t-tests on the $\maxodds$ showing significant differences between both the cross-language and control conditions ($p<0.05$).
    }
    \label{fig:summary}
\end{figure*}

\paragraph{Results} Our results are in \Cref{fig:summary}. Across layers, positions, and LMs, the cross-lingual generalization $\odds$ (green lines) are consistently higher than controls (grey lines), and at similar levels across phenomena. We also see the information flow through the LMs: early layers possess causally efficacious information at early positions, with the information passing through the middle layers in middle positions before reaching later layers at the final position. Pairwise t-tests with family-wise Holm-Bonferroni corrections show significantly higher transfer $\maxodds$ cross-lingually than to corresponding controls ($p<0.05$).\footnote{For bar charts of the $\maxodds$ see \Cref{app:max_odds_bars}} We see qualitatively similar results for \textsc{OOD labels}, with both transfer condition's $\maxodds$ broadly significantly above controls. This provides confidence that our results are not just overfit to the lexical items in our templates (full results in \Cref{appsec:ood_evals}). All told, these results provide evidence of shared, causally efficacious, syntactic subspaces in multilingual LMs, suggesting the emergence of language-agnostic mechanisms for these constructions. 

However, this generalization is not ubiquitous. \Cref{fig:summary} visually suggests transfer is highest within families (yellow lines are consistently above green lines), but pairwise t-tests on $\maxodds$ show mixed results. For number agreement, the within-family and all-language groups are significantly different at \textsc{PP} for all models, at \textsc{NP$_1$} for \texttt{tiny-aya-base} and \texttt{Llama-3.2-3b} and at \textsc{NP$_2$} for \texttt{Llama-3.2-3b}. For gender agreement, we find significant differences at \textsc{VP} and \textsc{name} for \texttt{mGPT}, \texttt{Llama-3.2-3b} and \texttt{tiny-aya-base}. Finally, for filler--gaps, we find no within-family vs. cross-language difference. This may reflect the low diversity of this language set or the high within-language $\odds$ and consistent cross-lingual transfer $\odds$ deflating the normalized $\maxodds$ we test with respect to the other constructions. 

\paragraph{Discussion} Our results show LMs converge to abstract cross-lingual syntactic mechanisms. Further, they suggest representational sharing is not without structure, but such structure is not as simple as, say, language family. However, our current experimental set-up does not lend insights into which factors are responsible for transfer. This question is centered in \Cref{sec:exp2}.

We further find evidence against the alternative, lexical-translation subspace hypothesis. That is, our results in \Cref{appsec:ood_evals} demonstrate interventions generalize across lexical items (both within a language and cross-lingually). These results provide evidence our results are not merely lexically specific, and therefore the subspaces not just translation-equivalent. We revisit this hypothesis in \Cref{sec:exp2}, where the typological structure of transfer further bears negative evidence for it.

\begin{figure}
    \centering
    \includegraphics[width=.95\linewidth]{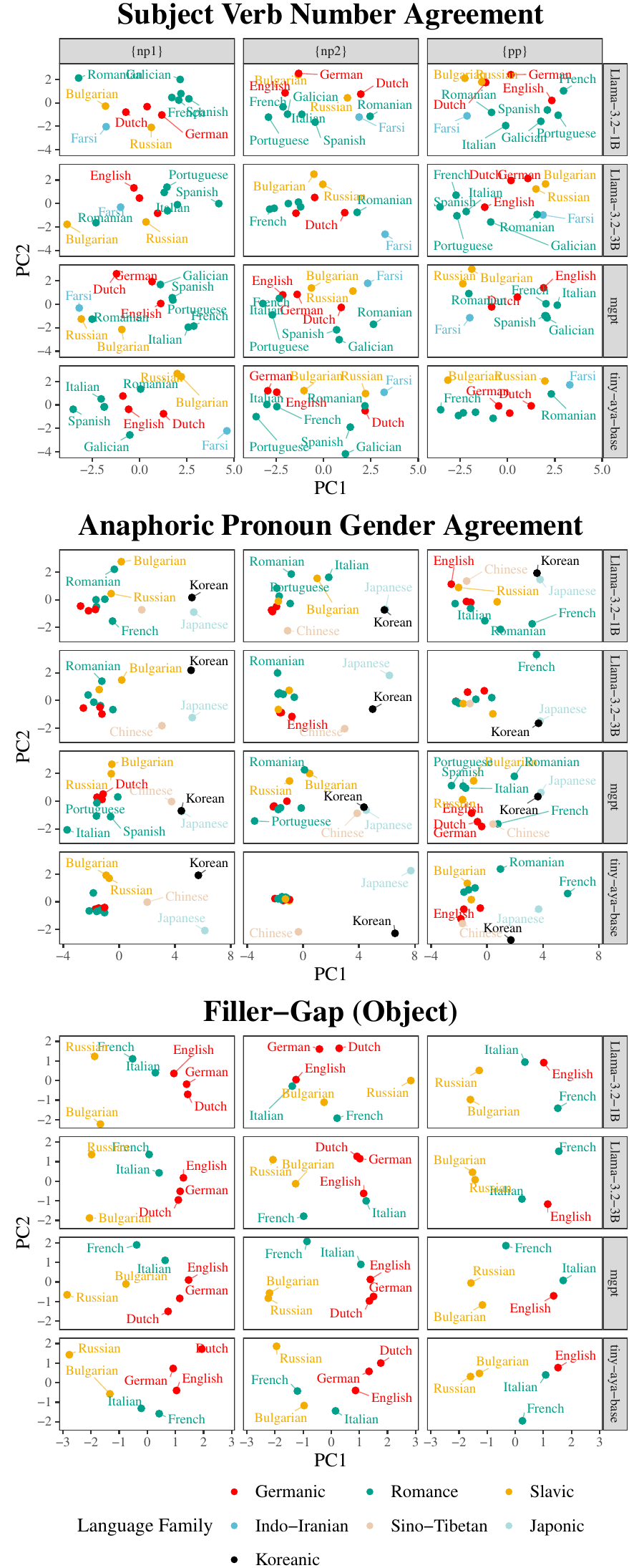}
    \caption{We visualize our languages on the top principal components of the transfer matrix. We find language family-based clusters not to be pure.}
    \label{fig:pca}
\end{figure}

\section{Experiment 2: What Drives Representational Reuse?}\label{sec:exp2}

We have found the syntactic mechanisms utilized by LMs can be transferred cross-lingually. However, exactly what governs this transfer is unclear. Furthermore, we have not teased apart the effects of the LM itself. Here, we analyze these factors.

\begin{table*}[ht]
  \centering
  \small
  \setlength{\tabcolsep}{4pt}
  \resizebox{\textwidth}{!}{%
  \begin{tabular}{l ccc ccc ccc}
  \toprule
   & \multicolumn{3}{c}{\textbf{Number Agreement}} & \multicolumn{3}{c}{\textbf{Gender Agreement}} & \multicolumn{3}{c}{\textbf{Filler--Gap (Object)}} \\
  \cmidrule(lr){2-4} \cmidrule(lr){5-7} \cmidrule(lr){8-10}
  \textbf{Term} & \textsc{NP$_1$} (.68) & \textsc{PP} (.71) & \textsc{NP$_2$} (.85) & \textsc{Name} (.69) & \textsc{VP} (.75) & \textsc{because} (.76) & \textsc{Filler} (.73) & \textsc{NP} (.61) & \textsc{Verb} (.78) \\
  \midrule
  Typological dist.
    & $\mathbf{-0.577}^{***}$
    & $\mathbf{-0.558}^{***}$
    & $\mathbf{-0.432}^{***}$
    & $\mathbf{-0.462}^{***}$
    & $\mathbf{-0.320}^{***}$
    & $\mathbf{-0.498}^{***}$
    & $\mathbf{-0.544}^{***}$
    & $\mathbf{-0.781}^{***}$
    & $\mathbf{-1.04}^{***}$ \\
  Model size
    & $0.363$
    & $0.268$
    & $0.481$
    & $0.400$
    & $0.379$
    & $0.550$
    & $-0.013$
    & $-0.151$
    & $0.224$ \\
  Typological dist.\ $\times$ model size
    & $-0.043$
    & $\mathbf{-0.100}^{***}$
    & $\mathbf{-0.078}^{**}$
    & $-0.055$
    & $\mathbf{-0.158}^{***}$
    & $-0.063$
    & $0.015$
    & $0.010$
    & $-0.103$ \\
  \bottomrule
  \end{tabular}
  }
  \caption{LMEM coefficients predicting the $\maxodds$ from linguistic and architectural factors. Estimates significant under the LRT are \textbf{bolded}. $^{*}p < .05$; $^{**}p < .01$; $^{***}p < .001$. $R^2_c$ values in parentheses next to each position. We find significant negative effects of typological distance, suggesting that transfer is stronger between typologically similar languages. Interaction effects between typological distance and model size further suggest such effects grow stronger in larger models for the tasks of Number Agreement and Gender Agreement.}
  \label{tab:pooled_lmem}
  \end{table*}

\paragraph{Set-Up} For each model, phenomenon, and position, we create an $n\times n$ matrix, $M$, where $n$ is the number of languages evaluated, and cell $M_{i,j}$ is the $\maxodds$ of the interventions trained on language $i$ when evaluated on language $j$. We then perform dimensionality reduction with PCA. Visualizing where our languages fall in this space provides insights about what languages have similar transfer patterns --- an implicit representation of which languages our LMs represent similarly. 

We also fit a linear mixed-effects model (LMEM) predicting the $\maxodds$ at each position from linguistic and LM-specific predictors. In particular, our LMEM has the LM, training language, and evaluation language as random effects with the typological distance between two languages, model size and their interaction as fixed effects. We operationalize typological distance as the cosine distance between \textsc{lang2vec} syntax vectors \citep{wals, littell-etal-2017-uriel}. For full LMEM details, see \Cref{appsec:lmem}.

\paragraph{Hypothesis}

We expect more transfer between typologically similar languages. We further expect smaller models to show more transfer as their smaller parameter count would explicitly promote representational reuse.

\paragraph{Results} Our PCA analysis (\Cref{fig:pca}) shows some clustering by language family: Romance, Slavic and Germanic languages generally cluster with themselves, with more typologically distant languages further afield. However, closer analysis suggests language family alone does not explain clustering: the average silhouette score (i.e., cluster purity) across language families with more than one language demonstrates low cohesion (\Cref{tab:silouhette}).

To decipher the role of other factors, we turn to our LMEM (\cref{tab:pooled_lmem}). Across all phenomena and positions, typological distance is a strong, significant predictor of transfer, with less distance between two languages associated with more transfer. That is, we consistently see stronger transfer between languages that are more syntactically similar --- independent of construction or model. We do not find a significant main effect of model size at any position --- refuting our hypothesis that smaller models would show more transfer. We do observe significant interactions of model size and typological distance at several positions across the agreement templates, with it consistently indicating that typological distance becomes a stronger predictor of transfer as a model gets larger, albeit weakly.

However, \textsc{lang2vec} is a coarse representation of a language's typological aggregate and does not tell us what lower-level features may be governing such transfer. We believe there are many factors which shape representational similarity during training, such as tokenizer overlap between languages, syntactic feature overlap, and number of cognates. In order to further characterize linguistic similarity, we correlate such metrics with each other, and the previously used typological distance, to show that such metrics generally covary. 

\begin{table}[ht!]
\centering
\footnotesize
\setlength{\tabcolsep}{4pt}
\resizebox{.7772\linewidth}{!}{%
\begin{tabular}{lrrr}
\toprule
\rowcolor{gray!30} \multicolumn{4}{l}{\textit{Subject Verb Number Agreement}} \\
& \textsc{NP$_1$} & \textsc{PP} & \textsc{NP$_2$}  \\
\midrule
\quad mGPT & -0.080 & 0.201 & -0.158 \\
\quad Llama-3.2-1b & 0.065 & 0.029 & -0.093 \\
\quad Llama-3.2-3b & -0.179 & 0.097 & 0.120 \\
\quad Tiny-Aya & 0.236 & 0.039 & -0.229 \\
\hline
\rowcolor{gray!30} \multicolumn{4}{l}{\textit{Anaphoric Pronoun Gender Agreement}} \\
& \textsc{Name} & \textsc{VP} & \textsc{because} \\
\midrule
\quad mGPT & 0.226 & -0.038 & -0.175 \\
\quad Llama-3.2-1b & -0.011 & 0.015 & -0.181 \\
\quad Llama-3.2-3b & -0.093 & 0.081 & -0.444\\
\quad Tiny-Aya & 0.087 & -0.104 & -0.322 \\
\hline
\rowcolor{gray!30} \multicolumn{4}{l}{\textit{Filler--Gap (Object)}} \\
& \textsc{Filler} & \textsc{NP} & \textsc{Verb} \\
\midrule
\quad mGPT & 0.387 & 0.543 & 0.106 \\
\quad Llama-3.2-1b & 0.220 & -0.019 & -0.003 \\
\quad Llama-3.2-3b & 0.297 & 0.232 & 0.259 \\
\quad Tiny-Aya & 0.331 & 0.134 & 0.277 \\
\bottomrule
\end{tabular}
}
\caption{Silhouette scores measuring language-family cluster purity in PCA space.
Each cell is the mean silhouette width for the given model and position.
Scores range from $-1$ (languages closer to other families) to $+1$ (tight within-family clusters).}
\label{tab:silouhette}
\end{table}

\begin{figure}
    \centering
    \includegraphics[width=\linewidth]{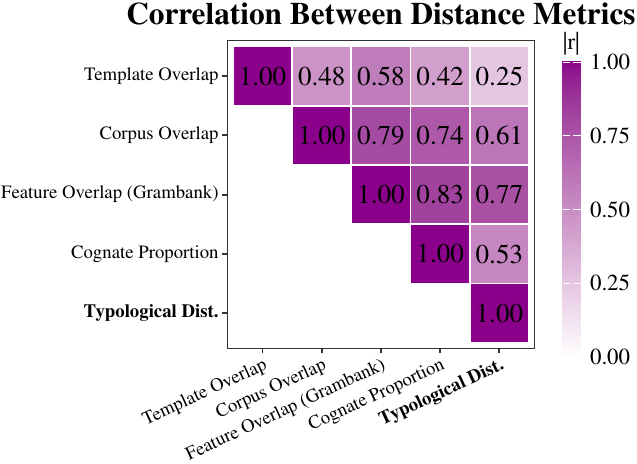}
    \caption{We measure the correlation between different operationalizations of typological similarity, pooled across positions and phenomenon. All pairwise correlations are significantly positive (all cells $p<.01$) and broadly strong, suggesting these factors are entangled in the model's training data and learned representations. Correlations with template overlap tend to be weaker, reflecting the hand-designed nature of our templates.}
    \label{fig:metrics}
\end{figure}

In particular, we take four measures of linguistic overlap: (1) \textbf{template-level tokenizer overlap}, calculated as the Jensen-Shannon divergence between token distributions at each position of our templates, pooled across our phenomenon and positions (2) \textbf{corpus-level tokenizer overlap}, calculated as the Jensen-Shannon divergence between token distributions across all text in the FLORES+ dataset \citep[a dataset consisting of 2019 sentences translated from English into 227 different languages;][]{nllb-24}, (3) \textbf{overlap in GramBank features}  \citep{skirgardGrambankRevealsImportance2023}, calculated as the Hamming Distance between binary feature vectors, and (4) \textbf{proportion of cognates}, calculated as the proportion of cognates in the LexiBank Indo-European Cognate Relationships database \citep{List2022Lexibank, Heggarty2024IECoR, Blum2025Lexibank2Description, Blum2026Lexibank2Data}. While there is broad coverage of these metrics across languages, we note there is not perfect coverage. Accordingly, we subset correlations to languages with such data.\footnote{Each metric's language coverage is in \Cref{appsec:coverage}.} 

Correlations are in \Cref{fig:metrics}. All pairwise correlations are significant, and generally strongly positive. Template overlap shows weaker correlation, likely due to our hand-designed templates differing from the overall language distribution and the low token variance of positions with fixed inventories, such as \textsc{filler} (2 items) and \textsc{because} (1 item). More broadly, these results suggest that typological similarity is broadly an effect of many lower-level features. Unfortunately, our data is not powered enough to perform regressions on individual features, leaving such analysis to future work.

\paragraph{Discussion} Our results show LM representational reuse is governed by high-level linguistic features, with mechanisms transferring  more between linguistically similar languages. We further demonstrate that \textsc{lang2vec} typological distance is strongly correlated with a broader set of features, both surface-level (tokenizer overlap) and linguistically driven (Grambank Feature overlap), which shape an LM's representational development during training. These results further provide evidence against the lexical-translation hypothesis because if our learned subspaces were merely lexical translations, we would not expect there to be typologically-graded cross-lingual transfer. 

Our results also bear on the debate regarding the role of tokenizer overlap in cross-lingual capabilities. Early work demonstrated overlap aids zero-shot transfer \citep{pires-etal-2019-multilingual}, but more recent work has suggested that language-specific tokenizers had benefit over shared ones \citep{rust-etal-2021-good}, and that overlap can even harm syntactic performance \citep{limisiewicz-etal-2023-tokenization}. More recently still, \citet{kallini-etal-2025-false} demonstrated that the semantic similarity of shared tokens, rather than overlap alone, drives cross-lingual benefit. While our results do not possess enough power to adjudicate on the debate, they suggest tokenizer overlap can be beneficial for cross-lingual mechanism sharing. However we also find above-control transfer between languages with disjoint scripts, suggesting tokenizer overlap alone is not the full story.

\section{Experiment 3: An Investigation Into the Effects of Training Data}

In their work on English filler--gap constructions, \citet{boguraev-etal-2025-causal} find that frequent constructions show more outward representational transfer, and infrequent constructions show more inward transfer. In our final experiment, we investigate whether there are similar effects of training data magnitude on multilingual representation sharing.

\paragraph{Set-Up} We focus on \texttt{mGPT} which is the only open-data model we study, trained on the multilingual Colossal Clean Crawled Corpus \citep[mC4;][]{JMLR:v21:20-074} and Wikipedia. Due to computational constraints, we do not rerun the full data generation pipeline, instead using per-language mC4 document counts as a proxy for training-data magnitude. To calculate inward and outward transfer, we follow \citet{boguraev-etal-2025-causal}: we compute the in- and out-degree of each language as a node in the transfer matrix across a sweep of edge-thresholds, with final inward and outward transfer values the area under the sweep's curve (AUC).

\paragraph{Hypothesis} We expect languages more prevalent in the training data to serve as stronger sources for transfer. Conversely, we expect less represented languages to serve as stronger sinks for transfer.

\paragraph{Results} Our results are in \cref{tab:correlations}. We consistently see a positive correlation between training-data and out-degree, but such a correlation is weak in two of our three phenomena. Further, only the filler--gap phenomenon shows a negative correlation between training-data and in-degree, with such correlation being negligible. Taken together, these results show little evidence for our hypothesis.

There is, however, strong correlation between a language's in- and out-degree AUCs (\cref{tab:correlations} third row). This result suggests multilingual transfer is not organized by `sources' and `sinks', but by `hubs' which promote and receive large amounts of transfer. We correlate a language's `hub'-ness (mean in- and out-degree AUC) with typological centrality (average pairwise typological distance across languages) and find the two to be highly correlated across phenomena (\cref{tab:correlations} fourth row). In comparison, `hub'-ness is generally not correlated with training data amount (\cref{tab:correlations} fifth row).

\begin{table}[ht!]
\centering
\resizebox{\columnwidth}{!}{%
\begin{tabular}{lrrr}
\toprule
 & \shortstack{Number\\Agreement} & \shortstack{Gender\\Agreement} & \shortstack{Filler--Gap\\(Object)} \\
\midrule
\rowcolor{gray!30}
\multicolumn{4}{l}{\textit{Training Data}} \\
\quad $r$(mC4, Out-Degree)  & 0.18 & 0.63 & 0.36  \\
\quad $r$(mC4, In-Degree)   & 0.25 & 0.77 & -0.06 \\
\hline
\rowcolor{gray!30}
\multicolumn{4}{l}{\textit{Typological `Hubs'}} \\
\quad $r$(In, Out)   & 0.70 & 0.89 & 0.44  \\
\quad $r$(Centr., Hub Score) & 0.59 & 0.85 & 0.75 \\
\quad $r$(mC4, Hub Score)  & 0.23 & 0.72 & 0.20  \\
\bottomrule
\end{tabular}%
}
\caption{Pearson correlation, $r$, between different data-metrics. `Hub'-ness is most correlated with typological centrality, not by training data magnitude.}
\label{tab:correlations}
\vspace{-10pt}
\end{table}

\paragraph{Discussion} Our results refute our hypothesis: transfer is not strongly coupled to training-data frequency. Instead, we find specific languages act as `hubs' in our transfer network, both receiving and promoting large amounts of representational transfer. We further find that `hub'-ness is strongly correlated with how similar a given language is to the other languages in the network. Such results provide additional evidence that multilingual representational sharing is governed by abstract linguistic features. We note these correlations are computed over few items ($n = 7\text{-}13$) and are therefore underpowered and suggestive, rather than conclusive.

\section{Conclusion} Linguists have long recognized seemingly parallel structure cross-lingually. These observations have led to many theories of cross-lingual regularities in linguistic mechanisms. However, thus far it has been hard to probe the degree to which internal processing overlaps in humans, and by which factors such overlap is governed. 

In this work, we take advantage of multilingual LMs to do just that. First, we show surface-level similarities are reflected in LM's internal organization, finding abstract mechanisms to be learned and repurposed by LMs cross-lingually. Second, we show such transfer is gradient, modulated by typological similarity between languages. Finally, we show there is little effect of training data magnitude on representation sharing, with transfer instead organized around typological `hubs'. 

Our findings point to linguistically interesting hypotheses regarding cross-linguistic syntactic structures and human multilingual processing --- hypotheses we imagine tested in psycholinguistic studies. For instance, syntactic priming paradigms could study whether languages showing greater transfer in LMs also exhibit stronger cross-linguistic priming in bilingual speakers. More broadly, we believe our work shows how the study of Language Models can help inform linguistic theory \citep{futrell_how_2025}.

\section*{Limitations}

This work demonstrates how Language Models can be used to study linguistically interesting questions. However the relationship between linguistic processing in neural models and humans is hotly debated \citep[e.g.,][\textit{i.a.}]{wilcox2020predictive, kuribayashi-etal-2021-lower, oh-schuler-2023-transformer, oh-schuler-2023-surprisal, oh-etal-2024-frequency, kuribayashi-etal-2025-large, piantadosi_2024_lms_refute_chomsky, katzir_why_2023, mahowald2024dissociating}. As such, our work should not be seen as providing definitive evidence about the nature of multilingual linguistic processing in humans, merely supporting evidence about the analyses general-purpose learners converge on to process such phenomena. 

Further, while DAS enables us to identify the presence of abstract cross-lingual mechanisms in a neural language model, we only have indirect access to the totality of the underlying mechanism via the design of the experimental stimuli: our results show that, e.g., nominal gender and verbal number features are causally related in the model's activations. DAS does not, however, reveal the complete mechanism the model uses to process the feature. For example, English marks only singular and plural, while Arabic also marks dual. Transferring an intervention on the number feature from English to Arabic might be disrupted by this extra dual category. Our method could reflect the disruption --- by comparing transfer quality across languages with and without a dual --- but it wouldn't be able to explicitly reveal dual number as part of the mechanism operating on number. As such, more work explicitly reverse-engineering complete mechanisms associated with linguistic features would be useful and elucidate this distinction. However, this is outside of our current scope.

Like much work on multilingual LMs, the range of languages we study here is a small subset of human languages and is skewed towards widely spoken ones. 
A more typologically diverse sample could potentially increase the richness of the conclusions we can draw.

Finally our work relies on templatically generated sentences to ensure large amounts of tight minimal pairs. Such sentences are known to differ from naturalistic sentences in meaningful ways. As such, extending this work to natural sentences would be a meaningful venture.


\section*{Acknowledgments}

We thank Olaf de Rohan Wilner, Agnese Lombardi, Youn-Gyu Park, Siyuan Song, Qing Yao, Leticia Hanada, Branimir Boguraev, Venus Shirazy, Maria Helena Fernandez Serrano, and Edoardo Giorgi for providing assistance with template design and judgments on resulting stimuli for the many different languages we study. We acknowledge funding from NSF CAREER grant 2339729 to Kyle Mahowald. Julius Steuer received funding from the Klaus Tschira Foundation, Heidelberg, Germany.

\bibliography{custom,anthology.min}

\appendix

\section{Templates}
\label{appsec:templates}

Exemplar stimuli we utilize to train and evaluate our interventions can be seen in \Cref{tab:data}. We also provide exemplar stimuli for our controls: exemplar stimuli for \textsc{random labels} can be seen in \Cref{tab:data-random}, and exemplar stimuli for \textsc{OOD labels} can be seen in \Cref{tab:data-ood}. We publish the full templates on Zenodo.\footnote{\href{https://doi.org/10.5281/zenodo.22025512}{Templates}}

\section{Intervention Models}

We publish all intervention models on Hugging Face.\footnote{
   \raisebox{-0.5pt}{\includegraphics[scale=0.036]{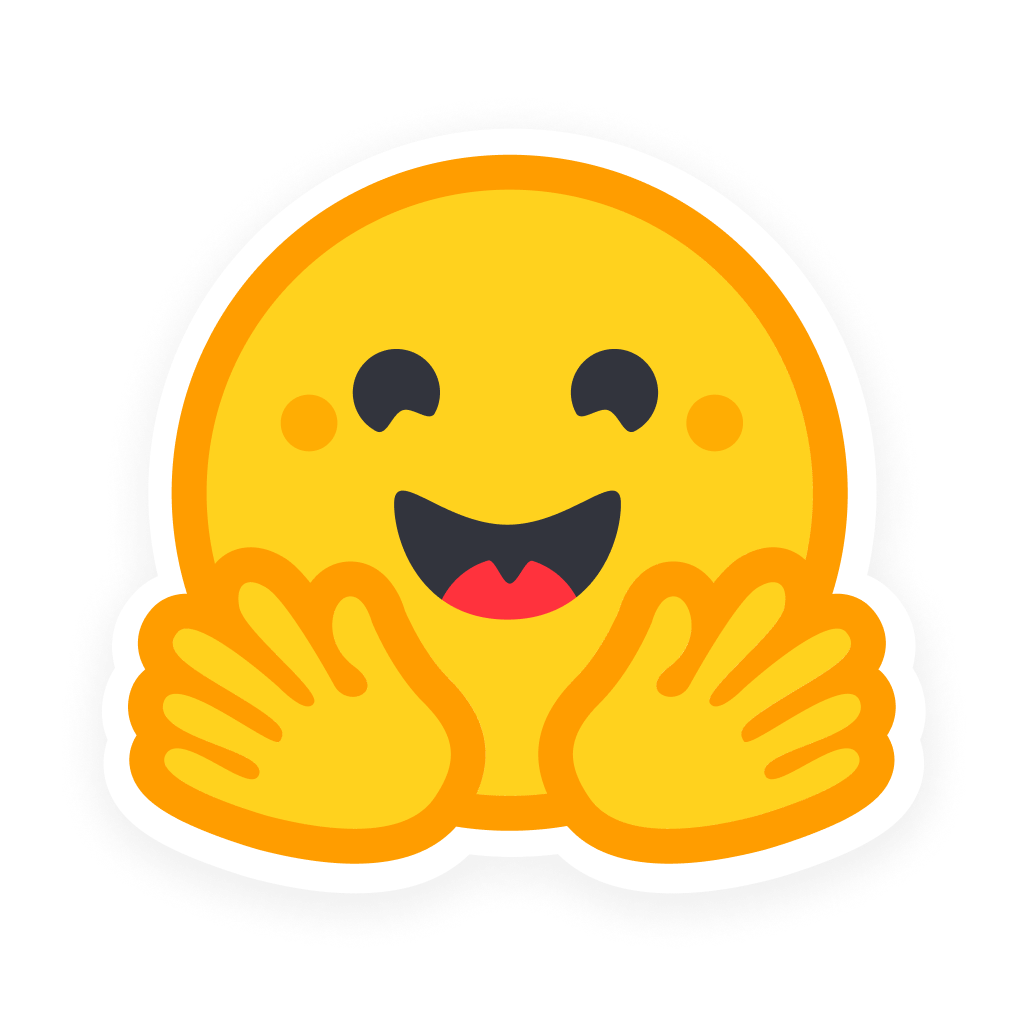} \href{https://huggingface.co/sashaboguraev/causal-multi-interventions}{Intervention models}}
}
The code used for model training, evaluation and analysis is available on GitHub.\footnote{\faGithub{} \href{https://github.com/justeuer/multilingual-interventions}{Training and evaluation code}}

\begin{table*}[!t]
\footnotesize
\centering

\begin{subtable}{\textwidth}
\centering
\setlength{\tabcolsep}{3pt}
\renewcommand{\arraystretch}{1.1}
\resizebox{\textwidth}{!}{%
\begin{tabularx}{\textwidth}{@{}p{1.8cm}XXXX@{}}
\toprule
\textsc{Language} & \textsc{NP$_1$} & \textsc{PP} & \textsc{NP$_2$} & \textsc{VP} \\
\midrule
English & The man / The men & near & the cabinet & is / are \\
Spanish & El hombre / Los hombres & cerca de & la alacena & es / son \\
German & Der Mann / Die Männer & neben & dem Schrank & ist / sind \\
Dutch & De man / De mannen & naast & de kast & is / zijn \\
French & L'homme / Les hommes & près de & la table & est / sont \\
Italian & L'uomo / Gli uomini & vicino alla & tavola & è / sono \\
Portuguese & O homem / Os homens & perto da & mesa & é / são \\
Romanian & Bărbatul / Bărbații & lângă & fereastră & este / sunt \\
Bulgarian & \cy{Мъжът / Мъжете} & \cy{до} & \cy{масата} & \cy{е / са} \\
Russian & \cy{Мужчина / Мужчины} & \cy{перед} & \cy{шкафом} & \cy{является / являются} \\
Farsi & \ar{مرد} / \ar{مردان} & \ar{کنار} & \ar{کمد} & \ar{است} / \ar{هستند} \\
Galician & O home / Os homes & preto & da fiestra & é / son \\
\bottomrule
\end{tabularx}}
\caption{Subject--Verb Number Agreement}
\label{tab:exemplars-sva-num}
\end{subtable}

\vspace{0.8em}

\begin{subtable}{\textwidth}
\centering
\setlength{\tabcolsep}{3pt}
\renewcommand{\arraystretch}{1.1}
\resizebox{\textwidth}{!}{%
\begin{tabularx}{\textwidth}{@{}p{1.8cm}XXXX@{}}
\toprule
\textsc{Language} & \textsc{Name} & \textsc{VP} & \textsc{Because} & \textsc{Pronoun} \\
\midrule
English & James / Mary & lied & because & he / she \\
Spanish & Miguel / María & corrió & porque & él / ella \\
German & Thomas / Maria & aß & weil & er / sie \\
Dutch & Jan / Anna & loog & omdat & hij / zij \\
French & Pierre / Marie & chuchota & parce qu' & il / elle \\
Italian & Mario / Maria & rise & perché & lui / lei \\
Portuguese & João / Maria & sussurra & porque & ele / ela \\
Romanian & Ion / Maria & a râs & pentru că & el / ea \\
Bulgarian & \cy{Иван / Мария} & \cy{вика} & \cy{защото} & \cy{той / тя} \\
Russian & \cy{Иван / Мария} & \cy{ходил / ходила} & \cy{потому что} & \cy{он / она} \\
Korean & \ko{철수 군이 / 지은 양이} & \ko{떨렸다,} & \ko{왜냐하면} & \ko{그가 / 그녀가} \\
Japanese & \jp{はるとくんが / さくらちゃんが} & \jp{来た、} & \jp{なぜなら} & \jp{彼が / 彼女が} \\
Chinese & \zh{何先生 / 袁夫人} & \zh{笑了,} & \zh{因为} & \zh{他 / 她} \\
\bottomrule
\end{tabularx}}
\caption{Anaphoric Pronoun Gender Agreement}
\label{tab:exemplars-sva-gender}
\end{subtable}

\vspace{0.8em}

\begin{subtable}{\textwidth}
\centering
\setlength{\tabcolsep}{3pt}
\renewcommand{\arraystretch}{1.1}
\resizebox{\textwidth}{!}{%
\begin{tabularx}{\textwidth}{@{}p{1.8cm}XXXXX@{}}
\toprule
\textsc{Language} & \textsc{Prefix} & \textsc{Filler} & \textsc{NP} & \textsc{VP} & \textsc{Gap} \\
\midrule
English & I know & that / what & she & saw & the / . \\
French & Je sais & qu' / ce qu' & elle & a vu & ça / . \\
German & Ich weiß, & dass / was & sie & --- & das / sah \\
Dutch & Ik weet & dat / wat & zij & --- & hem / zag \\
Italian & So & che / cosa & lei & vide & il / . \\
Russian & \cy{Я знаю,} & \cy{что / кого} & \cy{она} & \cy{видела} & \cy{это / .} \\
Bulgarian & \cy{Знам,} & \cy{че / какво} & \cy{тя} & \cy{видя} & \cy{него / .} \\
\bottomrule
\end{tabularx}}
\caption{Filler--Gap (Object).}
\label{tab:exemplars-fg_obj}
\end{subtable}

\caption{Exemplar minimal pairs per language for each construction. The minimal differences upstream (denoted by the two options separated by a `/') in the \textsc{NP$_1$}, \textsc{Name}, and \textsc{Filler} positions each correspond to the given labels in the \textsc{VP}, \textsc{Pronoun}, and \textsc{Gap} positions respectively. Such minimal pairs are used to evaluate LM linguistic competence, and to ensure that our causal interventions are successful. `-' indicates positions where the given construction has no item. This appears only in German and Dutch Filler--Gap stimuli as they are V2 languages, mandating any embedded verbs appear clause-final. The contrast at the VP position for Russian gender agreement (\cy{ходил}/\cy{ходила}) reflects the past-tense verb's obligatory agreement with the subject, and is not an additional label.}
\label{tab:data}
\end{table*}

\begin{table*}[!t]
\footnotesize
\centering

\begin{subtable}{\textwidth}
\centering
\setlength{\tabcolsep}{3pt}
\renewcommand{\arraystretch}{1.1}
\resizebox{\textwidth}{!}{%
\begin{tabularx}{\textwidth}{@{}p{1.8cm}XXXX@{}}
\toprule
\textsc{Language} & \textsc{NP$_1$} & \textsc{PP} & \textsc{NP$_2$} & \textsc{VP} \\
\midrule
English & The man / The men & near & the cabinet & dog / give \\
Spanish & El hombre / Los hombres & cerca de & la alacena & perro / dar \\
German & Der Mann / Die Männer & neben & dem Schrank & Hund / geben \\
Dutch & De man / De mannen & naast & de kast & hond / geven \\
French & L'homme / Les hommes & près de & la table & chien / donner \\
Italian & L'uomo / Gli uomini & vicino alla & tavola & cane / dare \\
Portuguese & O homem / Os homens & perto da & mesa & cão / dar \\
Romanian & Bărbatul / Bărbații & lângă & fereastră & câine / da \\
Bulgarian & \cy{Мъжът / Мъжете} & \cy{до} & \cy{масата} & \cy{куче / дава} \\
Russian & \cy{Мужчина / Мужчины} & \cy{перед} & \cy{шкафом} & \cy{собака / дать} \\
Farsi & \ar{مرد} / \ar{مردان} & \ar{کنار} & \ar{کمد} & \ar{سگ} / \ar{دادن} \\
Galician & O home / Os homes & preto & da fiestra & can / dar \\
\bottomrule
\end{tabularx}}
\caption{Subject--Verb Number Agreement (\textsc{random labels})}
\label{tab:exemplars-sva-num-random}
\end{subtable}

\vspace{0.8em}

\begin{subtable}{\textwidth}
\centering
\setlength{\tabcolsep}{3pt}
\renewcommand{\arraystretch}{1.1}
\resizebox{\textwidth}{!}{%
\begin{tabularx}{\textwidth}{@{}p{1.8cm}XXXX@{}}
\toprule
\textsc{Language} & \textsc{Name} & \textsc{VP} & \textsc{Because} & \textsc{Pronoun} \\
\midrule
English & James / Mary & lied & because & dog / give \\
Spanish & Miguel / María & corrió & porque & perro / dar \\
German & Thomas / Maria & aß & weil & Hund / geben \\
Dutch & Jan / Anna & loog & omdat & hond / geven \\
French & Pierre / Marie & chuchota & parce qu' & chien / donner \\
Italian & Mario / Maria & rise & perché & cane / dare \\
Portuguese & João / Maria & sussurra & porque & cão / dar \\
Romanian & Ion / Maria & a râs & pentru că & câine / da \\
Bulgarian & \cy{Иван / Мария} & \cy{вика} & \cy{защото} & \cy{куче / дава} \\
Russian & \cy{Иван / Мария} & \cy{ходил / ходила} & \cy{потому что} & \cy{собака / дать} \\
Korean & \ko{철수 군이 / 지은 양이} & \ko{떨렸다,} & \ko{왜냐하면} & \ko{개 / 주다} \\
Japanese & \jp{はるとくんが / さくらちゃんが} & \jp{来た、} & \jp{なぜなら} & \jp{犬 / あげる} \\
Chinese & \zh{何先生 / 袁夫人} & \zh{笑了,} & \zh{因为} & \zh{狗 / 给} \\
\bottomrule
\end{tabularx}}
\caption{Anaphoric Pronoun Gender Agreement (\textsc{random labels})}
\label{tab:exemplars-sva-gender-random}
\end{subtable}

\vspace{0.8em}

\begin{subtable}{\textwidth}
\centering
\setlength{\tabcolsep}{3pt}
\renewcommand{\arraystretch}{1.1}
\resizebox{\textwidth}{!}{%
\begin{tabularx}{\textwidth}{@{}p{1.8cm}XXXXX@{}}
\toprule
\textsc{Language} & \textsc{Prefix} & \textsc{Filler} & \textsc{NP} & \textsc{VP} & \textsc{Gap} \\
\midrule
English & I know & that / what & she & saw & dog / give \\
French & Je sais & qu' / ce qu' & elle & a vu & chien / donner \\
German & Ich weiß, & dass / was & sie & --- & Hund / geben \\
Dutch & Ik weet & dat / wat & zij & --- & hond / geven \\
Italian & So & che / cosa & lei & vide & cane / dare \\
Russian & \cy{Я знаю,} & \cy{что / кого} & \cy{она} & \cy{видела} & \cy{собака / дать} \\
Bulgarian & \cy{Знам,} & \cy{че / какво} & \cy{тя} & \cy{видя} & \cy{куче / дава} \\
\bottomrule
\end{tabularx}}
\caption{Filler--Gap (Object, \textsc{random labels}).}
\label{tab:exemplars-fg_obj-random}
\end{subtable}

\caption{Sampled exemplar minimal pairs per language for each construction for \textsc{random labels}: the stimulus is identical to the true construction, but the counterfactual target are arbitrary tokens. We expect a well-calibrated intervention to be \emph{specific} and not generalize to these.}
\label{tab:data-random}
\end{table*}

\begin{table*}[!t]
\footnotesize
\centering

\begin{subtable}{\textwidth}
\centering
\setlength{\tabcolsep}{3pt}
\renewcommand{\arraystretch}{1.1}
\resizebox{\textwidth}{!}{%
\begin{tabularx}{\textwidth}{@{}p{1.8cm}XXXX@{}}
\toprule
\textsc{Language} & \textsc{NP$_1$} & \textsc{PP} & \textsc{NP$_2$} & \textsc{VP} \\
\midrule
English & The man / The men & near & the cabinet & has / have \\
Spanish & El hombre / Los hombres & cerca de & la alacena & tiene / tienen \\
German & Der Mann / Die Männer & neben & dem Schrank & hat / haben \\
Dutch & De man / De mannen & naast & de kast & heeft / hebben \\
French & L'homme / Les hommes & près de & la table & a / ont \\
Italian & L'uomo / Gli uomini & vicino alla & tavola & ha / hanno \\
Portuguese & O homem / Os homens & perto da & mesa & tem / têm \\
Romanian & Bărbatul / Bărbații & lângă & fereastră & are / au \\
Bulgarian & \cy{Мъжът / Мъжете} & \cy{до} & \cy{масата} & \cy{може / могат} \\
Russian & \cy{Мужчина / Мужчины} & \cy{перед} & \cy{шкафом} & \cy{имеет / имеют} \\
Farsi & \ar{مرد} / \ar{مردان} & \ar{کنار} & \ar{کمد} & \ar{دارد} / \ar{دارند} \\
Galician & O home / Os homes & preto & da fiestra & está / están \\
\bottomrule
\end{tabularx}}
\caption{Subject--Verb Number Agreement (\textsc{OOD labels})}
\label{tab:exemplars-sva-num-ood}
\end{subtable}

\vspace{0.8em}

\begin{subtable}{\textwidth}
\centering
\setlength{\tabcolsep}{3pt}
\renewcommand{\arraystretch}{1.1}
\resizebox{\textwidth}{!}{%
\begin{tabularx}{\textwidth}{@{}p{1.8cm}XXXXX@{}}
\toprule
\textsc{Language} & \textsc{Prefix} & \textsc{Filler} & \textsc{NP} & \textsc{VP} & \textsc{Gap} \\
\midrule
English & I know & that / what & she & saw & a / , \\
French & Je sais & que / ce que & elle & a vu & cela / , \\
German & Ich weiß, & dass / was & sie & --- & das / fand \\
Dutch & Ik weet & dat / wat & zij & --- & hem / vond \\
Italian & So & che / cosa & lei & vide & un / , \\
Russian & \cy{Я знаю,} & \cy{что / кого} & \cy{она} & \cy{видела} & \cy{то / ,} \\
Bulgarian & \cy{Знам,} & \cy{че / какво} & \cy{тя} & \cy{видя} & \cy{това / ,} \\
\bottomrule
\end{tabularx}}
\caption{Filler--Gap (Object, \textsc{OOD labels}).}
\label{tab:exemplars-fg_obj-ood}
\end{subtable}

\caption{Sampled exemplar minimal pairs per language for each construction in the \textsc{OOD labels} control. As inspired by \citet{kumon-yanaka-2026-fine}: the stimulus is identical to the true construction, but the counterfactual target is a plausible yet out-of-distribution continuation. We expect a well-trained intervention to target the abstract feature of interest, and thus generalize to these.}
\label{tab:data-ood}
\end{table*}

\section{Behavioral Results}
\label{appsec:behavior}

\begin{figure*}
    \centering
    \includegraphics[width=\linewidth]{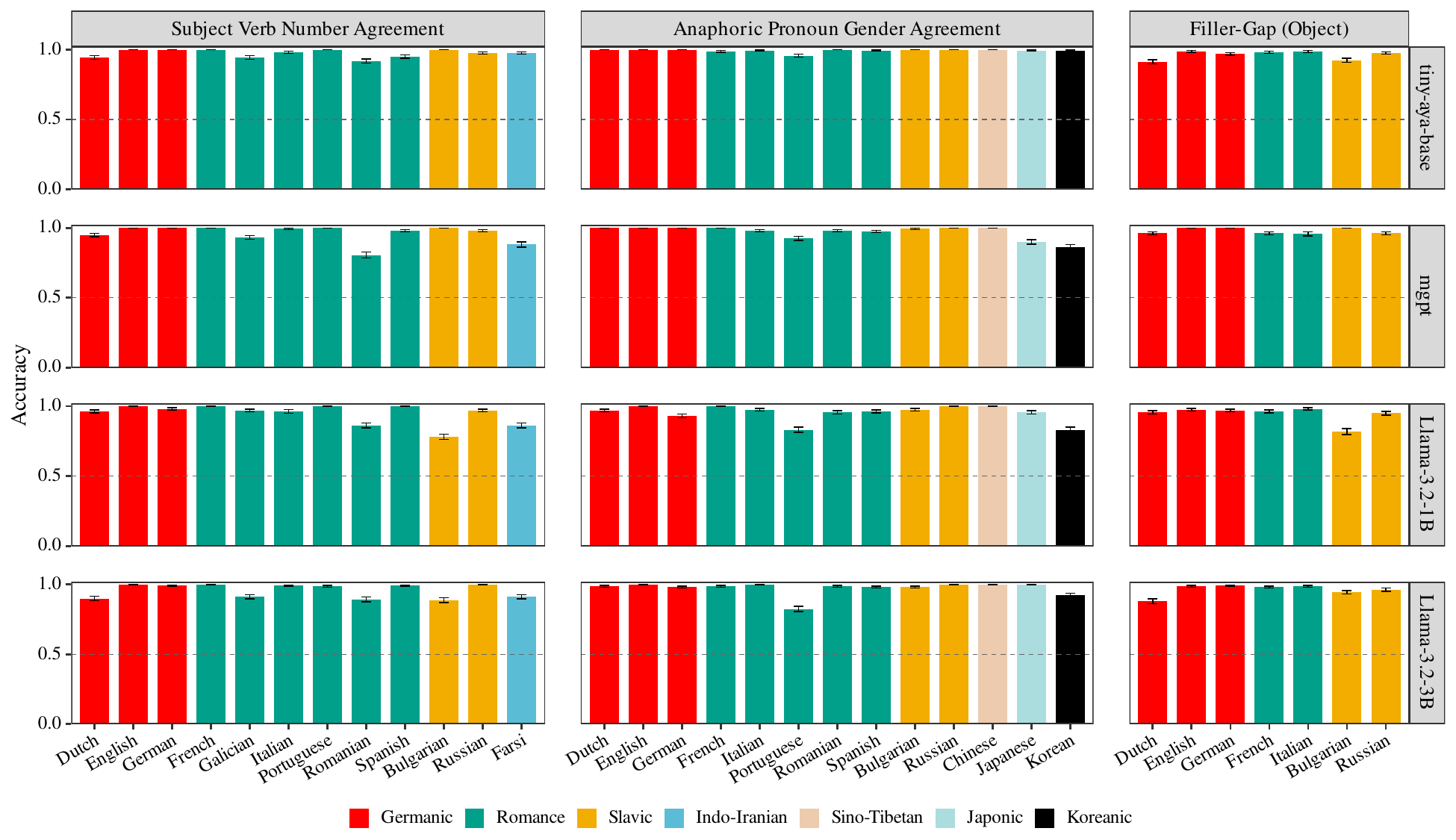}
    \caption{Behavioral performance of each model on each of our templatic stimuli. We broadly see strong performance from all of our models on all templates. Generally multilingual LMs perform the best, but both \texttt{Llama} models show strong performance as well. All bars but thirteen are above 90\% performance, with twelve of those above 80\%. The only bar below 80\% is \texttt{Llama‑3.2‑1b} on Bulgarian number agreement.}
    \label{fig:behavior}
\end{figure*}

Before attempting to study the internals of the LMs on our three phenomena, we must first confirm that they are behaviorally competent with the given task. We utilize a targeted syntactic evaluation paradigm inspired by the method developed by \citet{wilcox-etal-2018-rnn} to study RNN's competence of filler--gap constructions. The method is as follows: given two minimal pairs, $b$ and $s$, with corresponding labels $\ell_b$ and $\ell_s$, an LM should correctly predict that the surprisal, $S$, of each given label is higher given the correct context than in the incorrect context. Formally, the metric, termed \textsc{accuracy}, is a binary measure that is calculated as:

\[
\textsc{accuracy} =
\begin{cases}
1 & \text{if } S(\ell_b \mid b) < S(\ell_b \mid s) \\ & \text{ and } S(\ell_s \mid b) > S(\ell_s \mid s),\\[4pt]
0 & \text{otherwise.}
\end{cases}
\]
The LMs behavioral results can be seen in \Cref{fig:behavior}. Specifically, for each construction we study (column facets) and each LM (row facets), we show the mean-averaged performance over 160 randomly sampled minimal pairs. We can see that our LMs are broadly competent at processing such constructions, scoring at or above 90\% accuracy on 90\% (115/128) of templates and below 80\% accuracy only once (\texttt{Llama‑3.2‑1b} on Bulgarian number agreement). This gives us evidence that the LMs are syntactically competent with these phenomena, thus licensing our next experiments.

Digging deeper into these results, it is evident that the  expressly multilingual LMs are generally the most performant on the templates, perhaps as expected. Interestingly, however, \texttt{Llama-3.2-3b} does not lag far behind them. \texttt{Llama-3.2-1b} is the least performant of all the models on the task, however still showing strong performance. We do not see any systematic effects of language family, although we do note that the majority of these languages, and families, are high-resource and thus likely not data scarce during training.

\section{Max-Odds}\label{app:max_odds_bars}
\begin{figure*}
    \centering \includegraphics[width=0.95\linewidth]{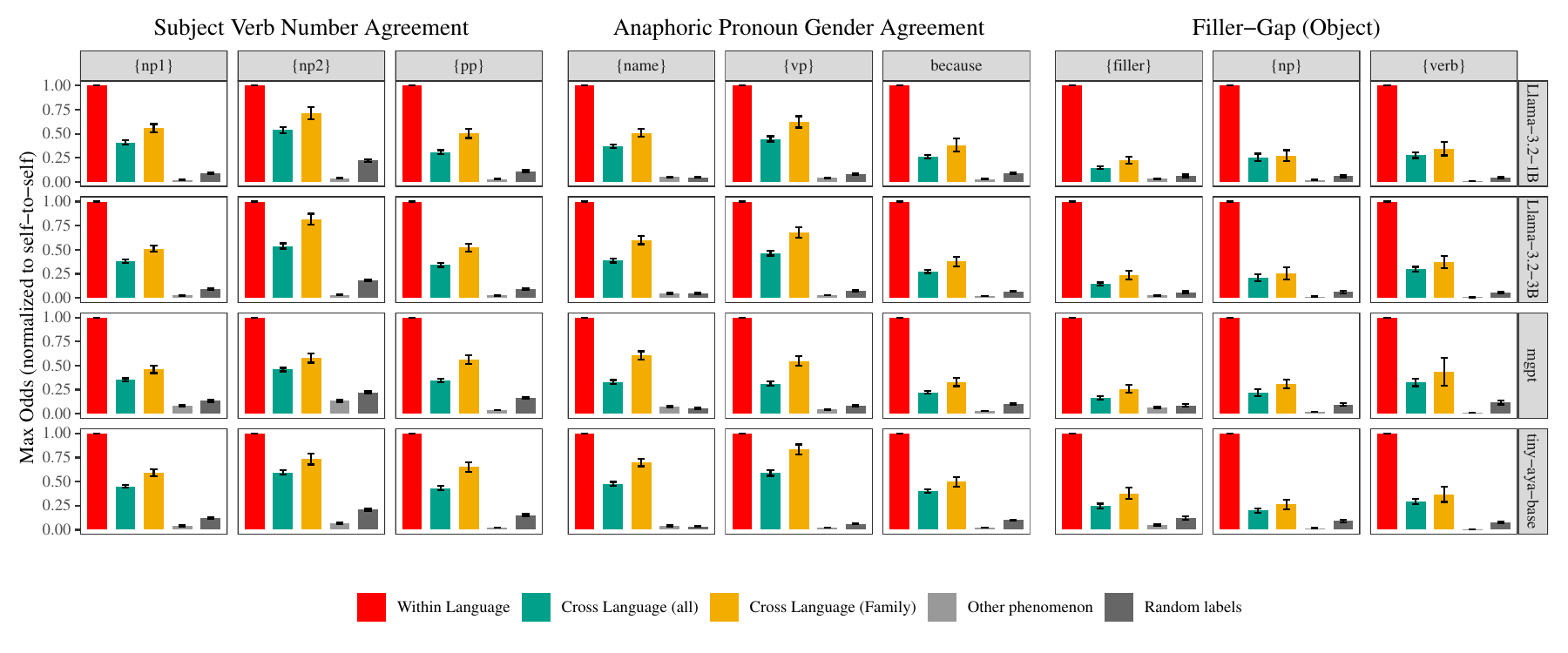}
    \caption{We take the $\maxodds$ across layers for each model at each position, before normalizing by the within-language value (red bar). We see strong performance in both our cross-language (green bar) and within-family (yellow bar) transfer conditions, with both being significantly above both controls in all cases when tested with t-tests with a Holm-Bonferroni correction. We further find the within-family transfer to be significantly higher than the cross-language bar in a select few comparisons, but for this to not hold more broadly.}
    \label{fig:bars}
\end{figure*}

The bar charts with the normalized $\maxodds$ for each of our five experimental groups in Experiment One can be seen in \Cref{fig:bars}. We further provide the Holm-Bonferroni corrected p-values of all reported comparisons for the main results in \Cref{tab:p-main}, and for the \textsc{OOD labels} in \Cref{tab:p-controls}.

\begin{table*}[ht]
\centering
\small
\setlength{\tabcolsep}{4pt}
\resizebox{\textwidth}{!}{%
\begin{tabular}{l ccccccccc}
\toprule
 & \multicolumn{3}{c}{\textbf{Filler--Gap (Object)}} & \multicolumn{3}{c}{\textbf{Subject Verb Number Agreement}} & \multicolumn{3}{c}{\textbf{Anaphoric Pronoun Gender Agreement}} \\
\cmidrule(lr){2-4} \cmidrule(lr){5-7} \cmidrule(lr){8-10}
\textbf{Model} & \textsc{Filler} & \textsc{NP} & \textsc{Verb} & \textsc{NP$_1$} & \textsc{PP} & \textsc{NP$_2$} & \textsc{NAME} & \textsc{VP} & \textsc{because} \\
\midrule
\rowcolor{gray!30}
\multicolumn{5}{l}{\textsc{Other phenomena}} &
\multicolumn{5}{r}{\textit{(\textbf{Control:} specificity)}} \\
\quad \texttt{Llama-3.2-1b} & $\mathbf{1.2\times10^{-9}}^{***}$ & $\mathbf{5.2\times10^{-7}}^{***}$ & $\mathbf{1.5\times10^{-7}}^{***}$ & $\mathbf{3.0\times10^{-40}}^{***}$ & $\mathbf{2.2\times10^{-28}}^{***}$ & $\mathbf{1.9\times10^{-34}}^{***}$ & $\mathbf{1.3\times10^{-32}}^{***}$ & $\mathbf{5.6\times10^{-30}}^{***}$ & $\mathbf{2.7\times10^{-22}}^{***}$ \\
\quad \texttt{Llama-3.2-3b} & $\mathbf{1.9\times10^{-7}}^{***}$ & $\mathbf{4.7\times10^{-6}}^{***}$ & $\mathbf{5.9\times10^{-9}}^{***}$ & $\mathbf{1.2\times10^{-41}}^{***}$ & $\mathbf{1.7\times10^{-35}}^{***}$ & $\mathbf{1.4\times10^{-40}}^{***}$ & $\mathbf{1.3\times10^{-31}}^{***}$ & $\mathbf{2.4\times10^{-35}}^{***}$ & $\mathbf{1.2\times10^{-33}}^{***}$ \\
\quad \texttt{mGPT} & $\mathbf{1.0\times10^{-5}}^{***}$ & $\mathbf{6.1\times10^{-6}}^{***}$ & $\mathbf{3.9\times10^{-7}}^{***}$ & $\mathbf{2.8\times10^{-35}}^{***}$ & $\mathbf{8.3\times10^{-36}}^{***}$ & $\mathbf{2.4\times10^{-33}}^{***}$ & $\mathbf{3.1\times10^{-22}}^{***}$ & $\mathbf{8.3\times10^{-25}}^{***}$ & $\mathbf{1.9\times10^{-32}}^{***}$ \\
\quad \texttt{tiny-aya-base} & $\mathbf{5.9\times10^{-9}}^{***}$ & $\mathbf{1.2\times10^{-9}}^{***}$ & $\mathbf{6.9\times10^{-9}}^{***}$ & $\mathbf{8.4\times10^{-58}}^{***}$ & $\mathbf{1.3\times10^{-39}}^{***}$ & $\mathbf{3.1\times10^{-52}}^{***}$ & $\mathbf{1.0\times10^{-44}}^{***}$ & $\mathbf{1.6\times10^{-40}}^{***}$ & $\mathbf{6.6\times10^{-49}}^{***}$ \\
\hline
\rowcolor{gray!30}
\multicolumn{5}{l}{\textsc{Random Labels}} &
\multicolumn{5}{r}{\textit{(\textbf{Control:} selectivity)}} \\
\quad \texttt{Llama-3.2-1b} & $\mathbf{2.0\times10^{-4}}^{***}$ & $\mathbf{3.9\times10^{-5}}^{***}$ & $\mathbf{1.3\times10^{-6}}^{***}$ & $\mathbf{1.7\times10^{-31}}^{***}$ & $\mathbf{8.0\times10^{-16}}^{***}$ & $\mathbf{5.9\times10^{-17}}^{***}$ & $\mathbf{4.1\times10^{-33}}^{***}$ & $\mathbf{2.2\times10^{-25}}^{***}$ & $\mathbf{1.0\times10^{-12}}^{***}$ \\
\quad \texttt{Llama-3.2-3b} & $\mathbf{4.1\times10^{-4}}^{***}$ & $\mathbf{5.4\times10^{-4}}^{***}$ & $\mathbf{7.4\times10^{-8}}^{***}$ & $\mathbf{7.0\times10^{-32}}^{***}$ & $\mathbf{4.1\times10^{-25}}^{***}$ & $\mathbf{1.4\times10^{-25}}^{***}$ & $\mathbf{1.2\times10^{-31}}^{***}$ & $\mathbf{8.7\times10^{-31}}^{***}$ & $\mathbf{3.1\times10^{-24}}^{***}$ \\
\quad \texttt{mGPT} & $\mathbf{0.002^{**}}$ & $\mathbf{0.004^{**}}$ & $\mathbf{2.6\times10^{-4}}^{***}$ & $\mathbf{1.8\times10^{-24}}^{***}$ & $\mathbf{7.6\times10^{-15}}^{***}$ & $\mathbf{7.0\times10^{-23}}^{***}$ & $\mathbf{1.8\times10^{-24}}^{***}$ & $\mathbf{6.7\times10^{-19}}^{***}$ & $\mathbf{1.7\times10^{-12}}^{***}$ \\
\quad \texttt{tiny-aya-base} & $\mathbf{4.0\times10^{-4}}^{***}$ & $\mathbf{4.1\times10^{-4}}^{***}$ & $\mathbf{5.4\times10^{-7}}^{***}$ & $\mathbf{1.1\times10^{-48}}^{***}$ & $\mathbf{3.3\times10^{-23}}^{***}$ & $\mathbf{2.3\times10^{-35}}^{***}$ & $\mathbf{6.6\times10^{-45}}^{***}$ & $\mathbf{2.6\times10^{-37}}^{***}$ & $\mathbf{1.2\times10^{-37}}^{***}$ \\
\hline
\rowcolor{gray!30}
\multicolumn{5}{l}{\textsc{Cross Language (Within-Family)}} &
\multicolumn{5}{r}{\textit{(\textbf{Test:} within-family vs.\ all-language transfer)}} \\
\quad \texttt{Llama-3.2-1b} & $0.843$ & $1.000$ & $1.000$ & $0.081^{\dagger}$ & $\mathbf{0.013^{*}}$ & $0.278$ & $0.066^{\dagger}$ & $0.208$ & $0.941$ \\
\quad \texttt{Llama-3.2-3b} & $0.843$ & $1.000$ & $1.000$ & $\mathbf{0.039^{*}}$ & $\mathbf{0.007^{**}}$ & $\mathbf{0.001^{**}}$ & $\mathbf{0.003^{**}}$ & $\mathbf{0.039^{*}}$ & $0.738$ \\
\quad \texttt{mGPT} & $0.843$ & $1.000$ & $1.000$ & $0.217$ & $\mathbf{0.002^{**}}$ & $0.422$ & $\mathbf{4.7\times10^{-5}}^{***}$ & $\mathbf{0.002^{**}}$ & $0.382$ \\
\quad \texttt{tiny-aya-base} & $0.799$ & $1.000$ & $1.000$ & $\mathbf{0.019^{*}}$ & $\mathbf{0.003^{**}}$ & $0.422$ & $\mathbf{1.9\times10^{-4}}^{***}$ & $\mathbf{0.007^{**}}$ & $0.843$ \\
\bottomrule
\end{tabular}
}
\caption{Holm-bonferroni corrected $p$-values (correction applied within each comparison type, pooled across phenomena/models/positions) for all pairwise $t$-tests comparing cross-lingual transfer (\textsc{Cross Language (all)}) against our two selectivity controls and against family-restricted transfer, for Experiment One (\Cref{sec:exp1};\Cref{fig:bars}). \textsc{Random Labels} and \textsc{Other phenomena} are our controls: cross-lingual transfer should, and does, significantly exceed both. \textsc{Cross Language (Within-Family)} instead tests whether transfer is stronger among typologically related languages than to the full evaluated set; results here are mixed. $^{\dagger}p<.1$; $^{*}p<.05$; $^{**}p<.01$; $^{***}p<.001$; significant cells bolded.}
\label{tab:p-main}
\end{table*}

\begin{table*}[ht]
\centering
\small
\setlength{\tabcolsep}{4pt}
\resizebox{\textwidth}{!}{%
\begin{tabular}{l cccccc}
\toprule
 & \multicolumn{3}{c}{\textbf{Filler--Gap (Object)}} & \multicolumn{3}{c}{\textbf{Subject Verb Number Agreement}} \\
\cmidrule(lr){2-4} \cmidrule(lr){5-7}
\textbf{Model} & \textsc{Filler} & \textsc{NP} & \textsc{Verb} & \textsc{NP$_1$} & \textsc{PP} & \textsc{NP$_2$} \\
\midrule
\rowcolor{gray!30}
\multicolumn{4}{l}{\textsc{Other phenomena}} &
\multicolumn{3}{r}{\textit{(\textbf{Control:} specificity)}} \\
\quad \texttt{Llama-3.2-1b} & $\mathbf{3.3\times10^{-7}}^{***}$ & $\mathbf{2.5\times10^{-6}}^{***}$ & $\mathbf{1.2\times10^{-6}}^{***}$ & $\mathbf{5.2\times10^{-39}}^{***}$ & $\mathbf{7.2\times10^{-32}}^{***}$ & $\mathbf{3.3\times10^{-40}}^{***}$ \\
\quad \texttt{Llama-3.2-3b} & $\mathbf{4.4\times10^{-6}}^{***}$ & $\mathbf{3.2\times10^{-6}}^{***}$ & $\mathbf{6.5\times10^{-7}}^{***}$ & $\mathbf{3.0\times10^{-42}}^{***}$ & $\mathbf{2.4\times10^{-40}}^{***}$ & $\mathbf{1.1\times10^{-43}}^{***}$ \\
\quad \texttt{mGPT} & $\mathbf{0.042^{*}}$ & $\mathbf{8.9\times10^{-6}}^{***}$ & $\mathbf{4.3\times10^{-6}}^{***}$ & $\mathbf{5.2\times10^{-30}}^{***}$ & $\mathbf{5.3\times10^{-44}}^{***}$ & $\mathbf{4.3\times10^{-55}}^{***}$ \\
\quad \texttt{tiny-aya-base} & $\mathbf{8.9\times10^{-6}}^{***}$ & $\mathbf{1.2\times10^{-8}}^{***}$ & $\mathbf{1.6\times10^{-9}}^{***}$ & $\mathbf{1.1\times10^{-51}}^{***}$ & $\mathbf{5.7\times10^{-45}}^{***}$ & $\mathbf{2.4\times10^{-59}}^{***}$ \\
\hline
\rowcolor{gray!30}
\multicolumn{4}{l}{\textsc{Random Labels}} &
\multicolumn{3}{r}{\textit{(\textbf{Control:} selectivity)}} \\
\quad \texttt{Llama-3.2-1b} & $0.427$ & $\mathbf{0.011^{*}}$ & $\mathbf{7.8\times10^{-5}}^{***}$ & $\mathbf{1.6\times10^{-31}}^{***}$ & $\mathbf{7.2\times10^{-19}}^{***}$ & $\mathbf{2.3\times10^{-22}}^{***}$ \\
\quad \texttt{Llama-3.2-3b} & $0.427$ & $0.427$ & $\mathbf{1.8\times10^{-4}}^{***}$ & $\mathbf{1.2\times10^{-32}}^{***}$ & $\mathbf{4.0\times10^{-30}}^{***}$ & $\mathbf{3.3\times10^{-28}}^{***}$ \\
\quad \texttt{mGPT} & $0.427$ & $0.488$ & $0.155$ & $\mathbf{1.2\times10^{-19}}^{***}$ & $\mathbf{6.3\times10^{-22}}^{***}$ & $\mathbf{2.9\times10^{-42}}^{***}$ \\
\quad \texttt{tiny-aya-base} & $0.327$ & $0.488$ & $\mathbf{3.4\times10^{-5}}^{***}$ & $\mathbf{8.7\times10^{-42}}^{***}$ & $\mathbf{8.9\times10^{-30}}^{***}$ & $\mathbf{7.6\times10^{-42}}^{***}$ \\
\hline
\rowcolor{gray!30}
\multicolumn{4}{l}{\textsc{Cross Language (Within-Family)}} &
\multicolumn{3}{r}{\textit{(\textbf{Test:} within-family vs.\ all-language transfer)}} \\
\quad \texttt{Llama-3.2-1b} & $1.000$ & $1.000$ & $1.000$ & $0.783$ & $0.083^{\dagger}$ & $0.442$ \\
\quad \texttt{Llama-3.2-3b} & $1.000$ & $1.000$ & $1.000$ & $0.304$ & $0.133$ & $0.148$ \\
\quad \texttt{mGPT} & $1.000$ & $0.839$ & $1.000$ & $0.304$ & $\mathbf{0.023^{*}}$ & $0.311$ \\
\quad \texttt{tiny-aya-base} & $1.000$ & $1.000$ & $1.000$ & $\mathbf{0.012^{*}}$ & $0.082^{\dagger}$ & $0.442$ \\
\bottomrule
\end{tabular}
}
\caption{Holm-bonferroni corrected $p$-values (correction applied within each comparison type, pooled across phenomena/models/positions) for the \textsc{OOD labels} (\Cref{appsec:ood_evals}; \Cref{fig:ood}). As in \Cref{tab:p-main}, \textsc{Random Labels} and \textsc{Other phenomena} are selectivity controls compared against \textsc{Cross Language (all)} transfer, and \textsc{Cross Language (Within-Family)} tests family-restricted vs.\ all-language transfer. $^{\dagger}p<.1$; $^{*}p<.05$; $^{**}p<.01$; $^{***}p<.001$; significant cells bolded.}
\label{tab:p-controls}
\end{table*}

\section{Out-of-Distribution Controls}\label{appsec:ood_evals}

As described in \Cref{sec:methods}, we also evaluate the generalization of our interventions to a set of lexically disjoint labels, which are still grammatically correct and exhibiting the same phenomenon we are testing. As noted above, we can only do this for two of our three phenomena --- subject--verb number agreement and filler--gap object extraction --- because there is not an alternative pronoun we can replace the `he/she' labels in our templates with robustly. 

\begin{figure*}
    \centering
    \begin{subfigure}{0.48\linewidth}
        \centering
        \includegraphics[width=\linewidth]{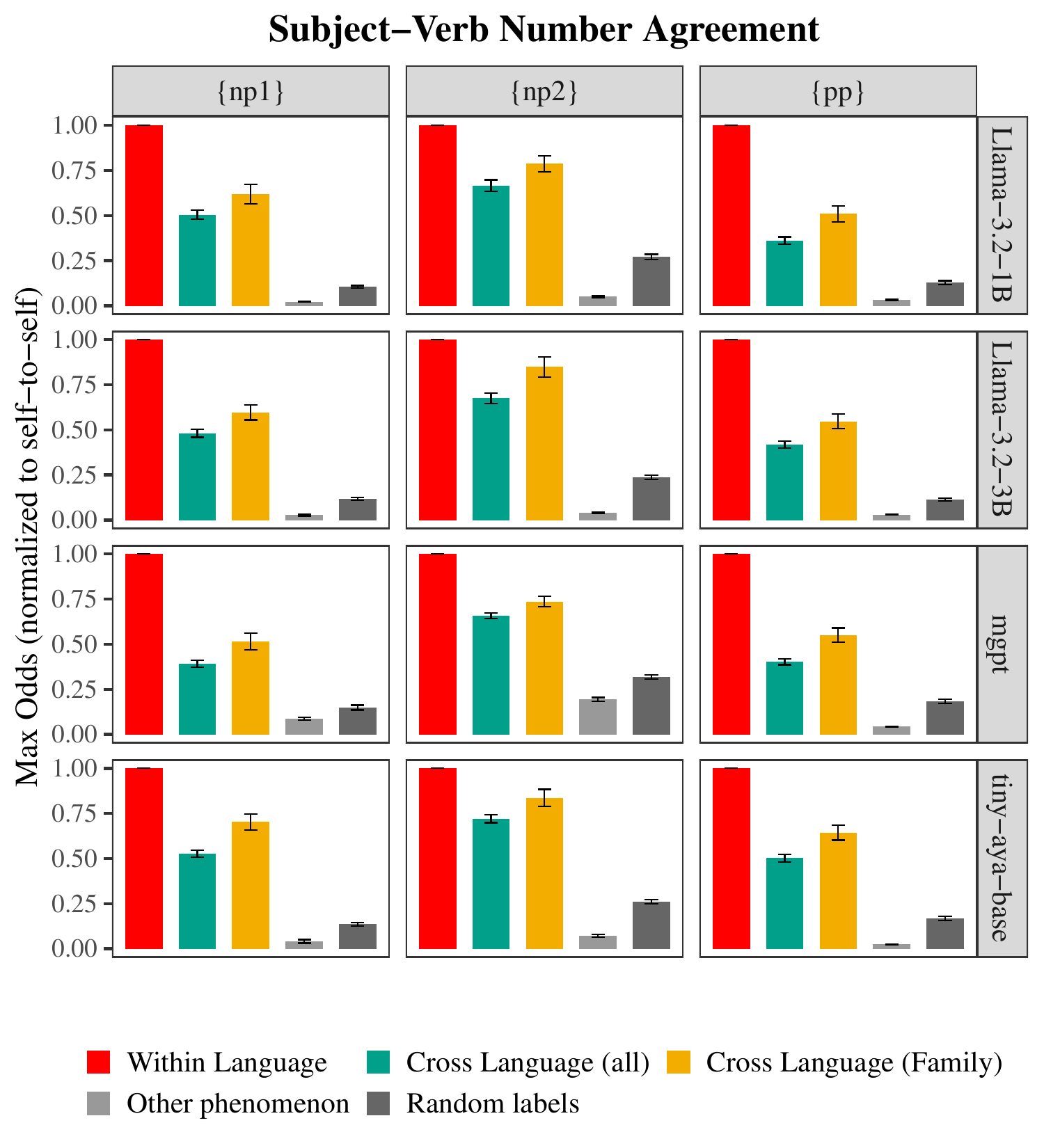}
        \label{fig:plot1}
    \end{subfigure}
    \hfill
    \begin{subfigure}{0.48\linewidth}
        \centering
        \includegraphics[width=\linewidth]{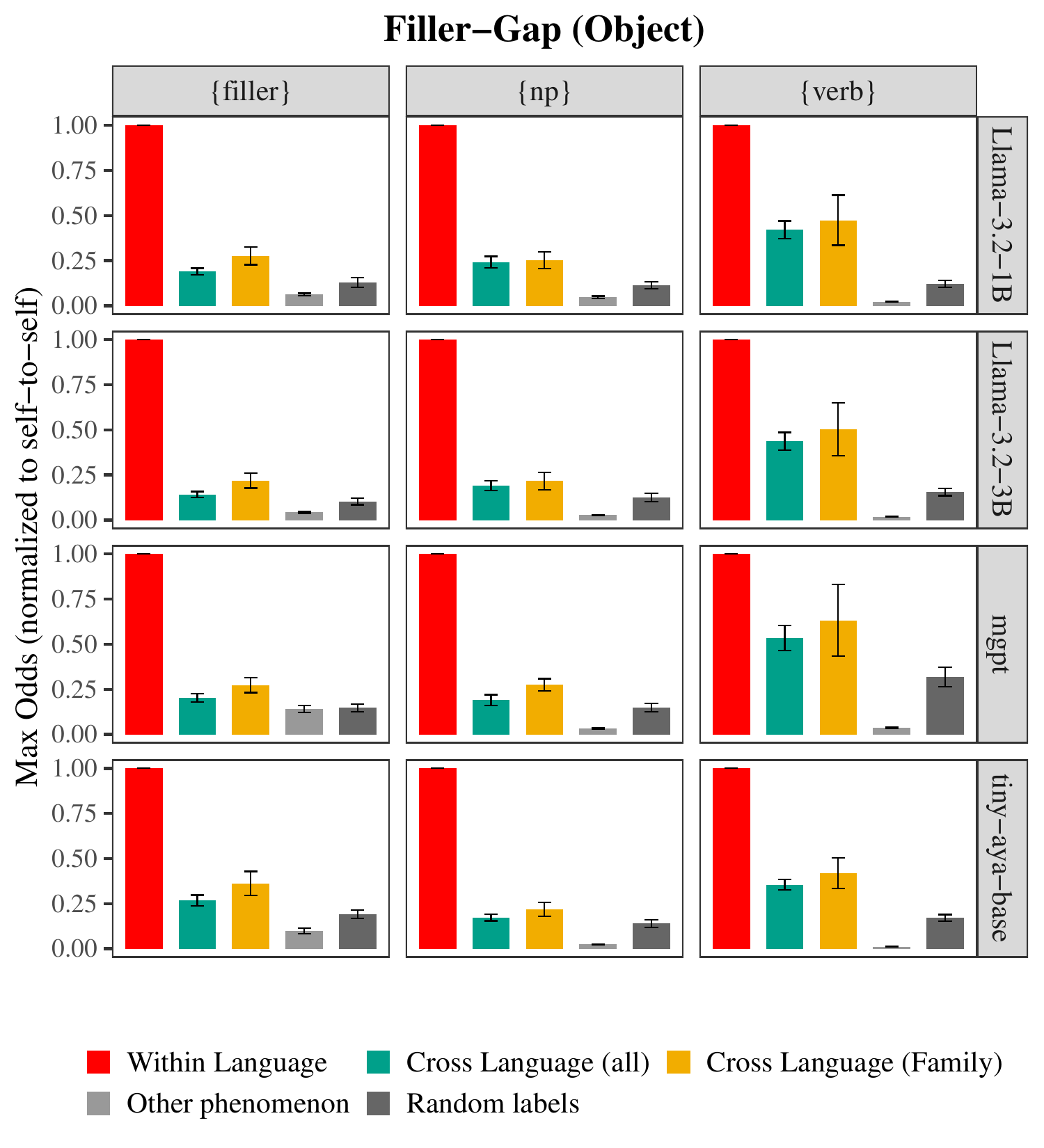}
        \label{fig:plot2}
    \end{subfigure}
    \caption{We evaluate our interventions on syntactically appropriate but lexically disjoint labels from our training set. We find our results to broadly hold in this setting.}
    \label{fig:ood}
\end{figure*}

Our results can be seen in \Cref{fig:ood}. Broadly, we see the same pattern as in our main set of results. Generalization across languages is consistently significantly higher than our across-phenomenon control for both phenomena, and consistently significantly higher than transfer to our random-label control in the case of subject--verb agreement. In the filler--gap case, we find generalization to OOD labels directionally correct for all comparisons. However, accompanying t-tests are broadly not significantly at most positions --- only showing significance at the \textsc{VP} position for both \texttt{Llama} models and \texttt{tiny-aya-base} and the \textsc{NP} position for \texttt{Llama-3.2-1b}. We attribute this to two main causes. First, there is a relatively lesser amount of samples for this phenomenon, which can make similar effect sizes less statistically significant, and secondly the raw transfer $\odds$ are similar cross-linguistically as in the other phenomena, but the raw within-language transfer is much higher. As such, the normalized $\maxodds$ are artificially deflated, leading to closer comparisons. These two effects likely cause the lack of significance we see. We also find there to be little difference between the unselective cross-lingual condition, and the family limited cross-lingual condition, with there being significant differences only at the \textsc{NP$_1$} position and \textsc{PP} position for \texttt{tiny-aya-base} and \texttt{mGPT} respectively. 

These results, in broadly matching what we find on our critical stimuli, provide us confidence that our interventions are not merely overfit to the lexical items we utilize in our templates, and instead indicative of general syntactic mechanisms we have uncovered.

\section{Linear-Mixed Effects Modeling Details}\label{appsec:lmem}

We perform all regressions with the \texttt{lmerTest} package in R \citep{kuznetsova2017lmertest}. We fit a linear mixed-effects model at each position with our dependent variable as the \textsc{\textbf{max odds}} at each training --- evaluation-set pair. We treat the LM, training language, and evaluation language as crossed random effects. Our fixed effects are the typological distance between the given languages, model size, and their interaction, with both scaled by z-scoring before regression. To estimate the $\beta$ coefficients, we fit our models with REML, and to estimate the significance of each term, we refit them with ML to calculate the likelihood-ratio test for each fixed-effect term against a nested model with that term removed. We fit the full model as per \citet{Barr2013RandomES}, namely with maximal random effects structure, removing collinear slopes until our model converges --- in this case this leaves us with an intercepts-only model. The final full model formula can be seen in \Cref{eq:lmem}.

\begin{equation}\label{eq:lmem}
\begin{aligned}
\text{max\_odds} \sim{} & \text{typological\_distance} * \text{model\_size} \\
& + (1 \mid \text{train\_lang}) \\
& + (1 \mid \text{eval\_lang}) \\
& + (1 \mid \text{model})
\end{aligned}
\end{equation}

\section{Metric Coverage}\label{appsec:coverage}

As mentioned in \Cref{sec:exp2}, the metrics we utilize to calculate typological distance do not have perfect coverage over our languages. We report the coverage in \Cref{tab:coverage}. Grambank does not have entries for German, Bulgarian, Spanish, and Romanian because it focuses on typological breadth and diversity. Furthermore, the cognate data we utilize, Cognate Proportion (IECOR) is limited to Indo-European languages, so it is unavailable for Galician, Chinese, Japanese, and Korean.

\begin{table*}[h]
\centering
\small
\begin{tabular}{l*{15}{c}}
\toprule
Metric &
\rotatebox{90}{English (en)} &
\rotatebox{90}{German (de)} &
\rotatebox{90}{Dutch (nl)} &
\rotatebox{90}{French (fr)} &
\rotatebox{90}{Italian (it)} &
\rotatebox{90}{Bulgarian (bg)} &
\rotatebox{90}{Russian (ru)} &
\rotatebox{90}{Spanish (es)} &
\rotatebox{90}{Portuguese (pt)} &
\rotatebox{90}{Romanian (ro)} &
\rotatebox{90}{Galician (gl)} &
\rotatebox{90}{Farsi (fa)} &
\rotatebox{90}{Chinese (zh)} &
\rotatebox{90}{Japanese (ja)} &
\rotatebox{90}{Korean (ko)} \\
\midrule
Template Overlap           & \yes & \yes & \yes & \yes & \yes & \yes & \yes & \yes & \yes & \yes & \yes & \yes & \yes & \yes & \yes \\
Corpus Overlap              & \yes & \yes & \yes & \yes & \yes & \yes & \yes & \yes & \yes & \yes & \yes & \yes & \yes & \yes & \yes \\
Feature Overlap (Grambank)  & \yes & \no  & \yes & \yes & \yes & \no  & \yes & \no  & \yes & \no  & \yes & \yes & \yes & \yes & \yes \\
Cognate Proportion          & \yes & \yes & \yes & \yes & \yes & \yes & \yes & \yes & \yes & \yes & \no  & \yes & \no  & \no  & \no  \\
Typological Dist.           & \yes & \yes & \yes & \yes & \yes & \yes & \yes & \yes & \yes & \yes & \yes & \yes & \yes & \yes & \yes \\
\bottomrule
\end{tabular}
\caption{Language coverage for each distance/overlap metric used in the typological-distance correlation analysis. Feature Overlap (Grambank) is missing for German, Bulgarian, Spanish, and Romanian because Grambank lacks entries for them. The cognate data we utilize, Cognate Proportion (IECOR) is limited to Indo-European languages, so it is unavailable for Galician, Chinese, Japanese, and Korean.}
\label{tab:coverage}
\end{table*}

\section{Computational Details}

We access all models studied through the \texttt{transformers} package \citet{wolf-etal-2020-transformers}, perform behavioral evaluations using the \texttt{minicons} package \citep{misra2022minicons} to compute surprisals and train DAS using the \texttt{pyvene} package \citep{wu-etal-2024-pyvene}. Training and evaluation is done using the same hyperparameters found in the extensive sweep performed by \citet{arora_causalgym_2024}. All experiments were ran on a cluster of 4 NVIDIA A40 (46 GB each, driver 610.43.02) GPUs, 64 CPU cores, 251 GB RAM, Linux/RHEL 8. Experiments were run as independent single-GPU jobs, up to 4 concurrently. Computational estimates on this hardware can be seen in \Cref{tab:compute}.

\begin{table*}[ht]
\centering
\small
\setlength{\tabcolsep}{5pt}
\resizebox{\textwidth}{!}{%
\begin{tabular}{l rrrr rrrr}
\toprule
 & \multicolumn{4}{c}{\textbf{Training}} & \multicolumn{4}{c}{\textbf{Evaluation}} \\
\cmidrule(lr){2-5} \cmidrule(lr){6-9}
\textbf{Model / Phenomenon} & Runs & \# Interventions & Seconds per Intervention & Time (h) & Pairs & \# Interventions & Seconds per Intervention & Time (h) \\
\midrule
\rowcolor{gray!30}
\multicolumn{9}{l}{\texttt{mGPT} (24 layers)} \\
\quad number agreement & 12 & 72 & 8.2 & 2.0 & 579 & 72 & 1.94 & 22.4 \\
\quad gender agreement & 13 & 72 & 8.2 & 2.1 & 534 & 72 & 2.83 & 30.3 \\
\quad filler--gap       &  7 & 72 & 8.1 & 1.1 & 294 & 72 & 5.14 & 30.2 \\
\quad \textbf{total}    & \textbf{32} & -- & -- & \textbf{5.3} & \textbf{1407} & -- & -- & \textbf{82.9} \\
\midrule
\rowcolor{gray!30}
\multicolumn{9}{l}{\texttt{tiny-aya-base} (36 layers)} \\
\quad number agreement & 12 & 108 & 16.9 & 6.1 & 579 & 108 & 3.09 & 53.6 \\
\quad gender agreement & 13 & 108 & 16.6 & 6.5 & 534 & 108 & 3.06 & 49.1 \\
\quad filler--gap       &  7 & 108 & 16.9 & 3.5 & 294 & 108 & 4.64 & 40.9 \\
\quad \textbf{total}    & \textbf{32} & -- & -- & \textbf{16.1} & \textbf{1407} & -- & -- & \textbf{143.6} \\
\midrule
\rowcolor{gray!30}
\multicolumn{9}{l}{\texttt{Llama-3.2-1b} (16 layers)} \\
\quad number agreement & 12 & 48 & 9.2  & 1.5 & 579 & 48 & 2.31 & 17.9 \\
\quad gender agreement & 13 & 48 & 10.6 & 1.8 & 534 & 48 & 3.22 & 22.9 \\
\quad filler--gap       &  7 & 48 & 9.5  & 0.9 & 294 & 48 & 5.50 & 21.6 \\
\quad \textbf{total}    & \textbf{32} & -- & -- & \textbf{4.2} & \textbf{1407} & -- & -- & \textbf{62.3} \\
\midrule
\rowcolor{gray!30}
\multicolumn{9}{l}{\texttt{Llama-3.2-3b} (28 layers)} \\
\quad number agreement & 12 & 84 & 16.0 & 4.5 & 579 & 84 & 2.99 & 40.4 \\
\quad gender agreement & 13 & 84 & 20.9 & 6.3 & 534 & 84 & 3.28 & 40.9 \\
\quad filler--gap       &  7 & 84 & 15.4 & 2.5 & 294 & 84 & 6.93 & 47.6 \\
\quad \textbf{total}    & \textbf{32} & -- & -- & \textbf{13.3} & \textbf{1407} & -- & -- & \textbf{128.9} \\
\midrule
\textbf{ALL} & \textbf{128} & -- & -- & $\approx$\textbf{39} & \textbf{5628} & -- & -- & $\approx$\textbf{418} \\
\bottomrule
\end{tabular}
}
\caption{Per-model compute cost on a single NVIDIA A40 GPU. A run is one language, with interventions at every position and layer. As there are three positions evaluated for each phenomenon, this just totals three times the number of layers in each model. Times reported reflect the median value of all runs performed during our experiments.}
\label{tab:compute}
\end{table*}

\section{LLM Usage Statement}

LLMs were utilized in the development of code and analyses, particularly for the adaptation of related codebases \citep[namely those of ][]{arora_causalgym_2024, boguraev-etal-2025-causal} to our work, as well as generating initial versions of some templates before refinement with native speaker informants. LLMs were further used to provide feedback on the clarity of our writing during the editing process.

\end{document}